\documentclass[11pt]{article}

\usepackage[utf8]{inputenc}
\usepackage[T1]{fontenc}
\usepackage{lmodern}

\usepackage{amsmath}
\usepackage{amssymb}
\usepackage{amsfonts}
\usepackage{mathtools}
\usepackage{amsthm}

\usepackage{graphicx}
\usepackage{booktabs}
\usepackage{array}
\usepackage{adjustbox}
\usepackage{enumitem}
\usepackage{microtype}
\usepackage{natbib}
\usepackage{xcolor}
\usepackage{url}

\usepackage{bm}

\usepackage{tabularx}
\usepackage{placeins}

\usepackage{tikz}
\usetikzlibrary{arrows.meta,positioning,calc,shapes.geometric,decorations.pathreplacing}
\usepackage{pgfplots}
\pgfplotsset{compat=1.18}
\usepackage{caption}
\usepackage{subcaption}

\usepackage{hyperref}

\usepackage{cleveref}

\graphicspath{{figures/}}
\newcolumntype{P}[1]{>{\raggedright\arraybackslash}p{#1}}

\newtheorem{theorem}{Theorem}
\newtheorem{proposition}[theorem]{Proposition}
\newtheorem{corollary}[theorem]{Corollary}

\theoremstyle{definition}
\newtheorem{definition}[theorem]{Definition}

\definecolor{DeckNavy}{HTML}{15324B}
\definecolor{DeckBlue}{HTML}{3B6EA8}
\definecolor{DeckTeal}{HTML}{2A8C82}
\definecolor{DeckOrange}{HTML}{E28A3B}
\definecolor{DeckRed}{HTML}{D9534F}
\definecolor{DeckGreen}{HTML}{59A14F}
\definecolor{DeckPurple}{HTML}{8C5B9E}
\definecolor{DeckGray}{HTML}{5D6770}

\theoremstyle{plain}

\newcommand{\R}{\mathbb{R}}

\newcommand{\tr}{\operatorname{tr}}
\newcommand{\diag}{\operatorname{diag}}

\newcommand{\rank}{\operatorname{rank}}

\newcommand{\norm}[1]{\lVert #1\rVert}
\newcommand{\ip}[2]{\langle #1,#2\rangle}
\newcommand{\Prob}{\mathbb P}
\newcommand{\calA}{\mathcal A}
\newcommand{\calC}{\mathcal C}
\newcommand{\calL}{\mathcal L}
\newcommand{\calR}{\mathcal R}
\newcommand{\calS}{\mathcal S}
\newcommand{\calT}{\mathcal T}

\newcommand{\spanop}{\operatorname{span}}

\newcommand{\col}{\operatorname{col}}
\newcommand{\range}{\operatorname{range}}
\newcommand{\nullsp}{\operatorname{null}}
\newcommand{\off}{\operatorname{off}}
\newcommand{\Var}{\operatorname{Var}}
\newcommand{\Cov}{\operatorname{Cov}}
\newcommand{\Id}{I}
\newcommand{\Mt}{\widetilde M}
\newcommand{\isafe}{i_{\mathrm{s}}}
\newcommand{\rinv}{r_{\mathrm{inv}}}
\newcommand{\rstar}{r^{\star}}
\newcommand{\rlam}{r_{\lambda}}
\newcommand{\E}{\mathbb{E}}

\newcommand{\by}[1]{\hfill\textup{(#1)}}
\newenvironment{claimproof}%
  {\begin{proof}\leavevmode\begin{enumerate}[label=\textup{(\arabic*)},leftmargin=*,itemsep=1pt,topsep=2pt]}%
  {\end{enumerate}\end{proof}}

\IfFileExists{results_macros.tex}{
\newcommand{\NumFibers}{8000}
\newcommand{\DimD}{64}
\newcommand{\NumFeatures}{192}
\newcommand{\LangBlockSize}{48}
\newcommand{\NoiseEta}{0.015}
\newcommand{\MaxExactGramError}{\ensuremath{2.66\times 10^{-15}}}
\newcommand{\MeanAbsNoiseResidual}{0.0012}
\newcommand{\NoiseResidualBound}{0.030}
\newcommand{\ResidualRatioTail}{0.106}
\newcommand{\DecisionDisagreementRate}{49.3}
\newcommand{\DirectedFailureRate}{31.4}
\newcommand{\ReverseFailureRate}{32.7}
\newcommand{\ConditionalDirectedFailureRate}{46.6}
\newcommand{\AucOracleDrift}{0.966}
\newcommand{\AucHiddenDistance}{0.707}
\newcommand{\AucDiagonalProxy}{0.588}
\newcommand{\AucPermutation}{0.506}
\newcommand{\MarginStrataTotal}{5}
\newcommand{\GramStrataOrdered}{5}
\newcommand{\PermutedStrataOrdered}{2}
\newcommand{\BlockLevel}{3.00}
\newcommand{\ProtectedRhoFeature}{0.000}
\newcommand{\ProtectedKappaHidden}{0.000}
\newcommand{\ProtectedDriftSd}{0.000}
\newcommand{\ProtectedDirectedFailurePct}{0.000}
\newcommand{\RandomRhoFeature}{0.899}
\newcommand{\RandomKappaHidden}{0.876}
\newcommand{\RandomDriftSd}{0.318}
\newcommand{\RandomDirectedFailurePct}{1.472}
\newcommand{\SpikedRhoFeature}{3.464}
\newcommand{\SpikedKappaHidden}{0.991}
\newcommand{\SpikedDriftSd}{1.083}
\newcommand{\SpikedDirectedFailurePct}{30.721}
\newcommand{\RandomIsotropicBenchmark}{0.866}
\newcommand{\RandomCenterMargin}{0.350}
\newcommand{\RandomStandardizedMargin}{2.203}
\newcommand{\RandomGaussianFailurePct}{1.38}
\newcommand{\SpikedCenterMargin}{0.350}
\newcommand{\SpikedStandardizedMargin}{0.646}
\newcommand{\SpikedGaussianFailurePct}{25.91}
\newcommand{\CommDistinctPairs}{48}
\newcommand{\CommInterventions}{768}
\newcommand{\CommIdentityError}{\ensuremath{4.41\times 10^{-16}}}
\newcommand{\CommCorr}{0.999}
\newcommand{\CommRmse}{0.0034}
\newcommand{\BlockRhoTrue}{1.933}
\newcommand{\BlockRhoEstimate}{1.939}

\newcommand{\SignedCommCorr}{0.9998}
\newcommand{\SignedCommRmse}{0.0050}
\newcommand{\BenignOverRefusalEn}{18.71}

\newcommand{\RepairSelectedLambda}{3.00}
\newcommand{\RepairBaseTestFailurePct}{0.506}
\newcommand{\RepairSelectedTestFailurePct}{0.164}
\newcommand{\RepairBaseTestDrift}{0.219}
\newcommand{\RepairSelectedTestDrift}{0.026}
\newcommand{\RepairSelectedTestWilsonLow}{0.084}
\newcommand{\RepairSelectedTestWilsonHigh}{0.303}
\newcommand{\TiedTestFailurePct}{28.79}
\newcommand{\OracleProjTestFailurePct}{0.04}
\newcommand{\TruncProjTestFailurePct}{2.16}
\newcommand{\TiedTestDrift}{1.394}
\newcommand{\OracleProjTestDrift}{0.000}
\newcommand{\TruncProjTestDrift}{0.008}
\newcommand{\OracleProjHeadNorm}{0.145}
\newcommand{\OracleRemovedHeadEnergy}{97.8}
\newcommand{\TruncBasisRank}{43.1}
\newcommand{\TiedFixedTauFailurePct}{33.15}
\newcommand{\OracleFixedTauFailurePct}{0.00}
\newcommand{\TruncFixedTauFailurePct}{0.00}
\newcommand{\MatchedSeeds}{12}
\newcommand{\MatchedFibers}{8000}
\newcommand{\MatchedRho}{0.880}
\newcommand{\StableChi}{0.000}
\newcommand{\IllChi}{0.000}
\newcommand{\IntrinsicChi}{0.880}
\newcommand{\StableInvariantNorm}{1.35}
\newcommand{\IllInvariantNorm}{91.59}
\newcommand{\StableDeployedDrift}{0.518}
\newcommand{\StableSelectedDrift}{0.150}
\newcommand{\IllDeployedDrift}{0.518}
\newcommand{\IllSelectedDrift}{0.453}
\newcommand{\IntrinsicDeployedDrift}{0.518}
\newcommand{\IntrinsicSelectedDrift}{0.519}
\newcommand{\StableDriftReduction}{71.0}
\newcommand{\IllDriftReduction}{12.7}
\newcommand{\IntrinsicDriftReduction}{-0.0}
\newcommand{\StableSelectedBacc}{99.99}
\newcommand{\IllSelectedBacc}{99.31}
\newcommand{\IntrinsicSelectedBacc}{98.76}
\newcommand{\StableExactBacc}{99.99}
\newcommand{\IllExactBacc}{52.16}

\newcommand{\DeployedScoreMismatch}{\ensuremath{2.22\times 10^{-15}}}
\newcommand{\SweepGeometries}{66}
\newcommand{\IntrinsicSweepSpearman}{0.972}
\newcommand{\ConditioningSweepSpearman}{0.889}
\newcommand{\IntrinsicLowResidualRatio}{0.494}
\newcommand{\IntrinsicHighResidualRatio}{1.000}
\newcommand{\StableSweepResidualRatio}{0.386}
\newcommand{\IllSweepResidualRatio}{0.885}

\newcommand{\LearningMinTrain}{16}
\newcommand{\LearningMaxTrain}{2048}
\newcommand{\StableRatioSixteen}{0.526}
\newcommand{\StableRatioMax}{0.290}
\newcommand{\IllRatioSixteen}{0.948}
\newcommand{\IllRatioMax}{0.872}
\newcommand{\IntrinsicRatioMax}{1.000}
\newcommand{\UntiedSystems}{8}
\newcommand{\UntiedTriples}{360}
\newcommand{\UntiedMedianResidualSpearman}{0.964}
\newcommand{\UntiedMedianCrossGramSpearman}{0.269}
\newcommand{\UntiedPooledResidualSpearman}{0.970}
\newcommand{\UntiedPooledCrossGramSpearman}{0.298}

\newcommand{\GlobalEqualRatio}{1.000}
\newcommand{\GlobalRandomOptimizedRatio}{1.022}
\newcommand{\GlobalRandomTiedRatio}{1.552}
\newcommand{\GlobalLeverageSpearman}{0.752}

\newcommand{\BudgetStableKappa}{0.670}
\newcommand{\BudgetIllKappa}{1.000}
\newcommand{\BudgetIntrinsicKappa}{1.000}
\newcommand{\ImportanceModels}{24}
\newcommand{\ImportanceSteps}{6000}
\newcommand{\ImportanceKept}{19}
\newcommand{\ImportanceTauUniformMin}{1.06}
\newcommand{\ImportanceTauUniformMax}{1.22}
\newcommand{\ImportanceTauNonuniformMax}{1.90}
\newcommand{\ImportanceAlignUniform}{0.992}
\newcommand{\ImportanceAlignNonuniform}{0.752}
\newcommand{\ImportanceRhoRatioUniform}{1.00}
\newcommand{\ImportanceRhoRatioNonuniform}{0.248}
\newcommand{\ImportanceRhoTopNonuniformMin}{\ensuremath{5.63\times 10^{-4}}}
\newcommand{\ImportanceRsqLevelUniform}{0.82}
\newcommand{\ImportanceRsqLevelPooled}{0.11}
\newcommand{\ImportanceRsqStructurePooled}{0.19}
\newcommand{\ImportanceWelchGapUniform}{0.026}
\newcommand{\ImportanceWelchGapNonuniform}{0.144}
}{
\providecommand{\NumFibers}{n/a}
\providecommand{\DimD}{n/a}
\providecommand{\NumFeatures}{n/a}
\providecommand{\LangBlockSize}{n/a}
\providecommand{\NoiseEta}{n/a}
\providecommand{\MaxExactGramError}{n/a}
\providecommand{\MeanAbsNoiseResidual}{n/a}
\providecommand{\NoiseResidualBound}{n/a}
\providecommand{\ResidualRatioTail}{n/a}
\providecommand{\DecisionDisagreementRate}{n/a}
\providecommand{\DirectedFailureRate}{n/a}
\providecommand{\ReverseFailureRate}{n/a}
\providecommand{\ConditionalDirectedFailureRate}{n/a}
\providecommand{\AucOracleDrift}{n/a}
\providecommand{\AucHiddenDistance}{n/a}
\providecommand{\AucDiagonalProxy}{n/a}
\providecommand{\AucPermutation}{n/a}
\providecommand{\MarginStrataTotal}{n/a}
\providecommand{\GramStrataOrdered}{n/a}
\providecommand{\PermutedStrataOrdered}{n/a}
\providecommand{\BlockLevel}{n/a}
\providecommand{\ProtectedRhoFeature}{n/a}
\providecommand{\ProtectedKappaHidden}{n/a}
\providecommand{\ProtectedDriftSd}{n/a}
\providecommand{\ProtectedDirectedFailurePct}{n/a}
\providecommand{\RandomRhoFeature}{n/a}
\providecommand{\RandomKappaHidden}{n/a}
\providecommand{\RandomDriftSd}{n/a}
\providecommand{\RandomDirectedFailurePct}{n/a}
\providecommand{\SpikedRhoFeature}{n/a}
\providecommand{\SpikedKappaHidden}{n/a}
\providecommand{\SpikedDriftSd}{n/a}
\providecommand{\SpikedDirectedFailurePct}{n/a}
\providecommand{\RandomIsotropicBenchmark}{n/a}
\providecommand{\RandomCenterMargin}{n/a}
\providecommand{\RandomStandardizedMargin}{n/a}
\providecommand{\RandomGaussianFailurePct}{n/a}
\providecommand{\SpikedCenterMargin}{n/a}
\providecommand{\SpikedStandardizedMargin}{n/a}
\providecommand{\SpikedGaussianFailurePct}{n/a}
\providecommand{\CommDistinctPairs}{n/a}
\providecommand{\CommInterventions}{n/a}
\providecommand{\CommIdentityError}{n/a}
\providecommand{\CommCorr}{n/a}
\providecommand{\CommRmse}{n/a}
\providecommand{\BlockRhoTrue}{n/a}
\providecommand{\BlockRhoEstimate}{n/a}

\providecommand{\SignedCommCorr}{n/a}
\providecommand{\SignedCommRmse}{n/a}
\providecommand{\BenignOverRefusalEn}{n/a}

\providecommand{\RepairSelectedLambda}{n/a}
\providecommand{\RepairBaseTestFailurePct}{n/a}
\providecommand{\RepairSelectedTestFailurePct}{n/a}
\providecommand{\RepairBaseTestDrift}{n/a}
\providecommand{\RepairSelectedTestDrift}{n/a}
\providecommand{\RepairSelectedTestWilsonLow}{n/a}
\providecommand{\RepairSelectedTestWilsonHigh}{n/a}
\providecommand{\TiedTestFailurePct}{n/a}
\providecommand{\OracleProjTestFailurePct}{n/a}
\providecommand{\TruncProjTestFailurePct}{n/a}
\providecommand{\TiedTestDrift}{n/a}
\providecommand{\OracleProjTestDrift}{n/a}
\providecommand{\TruncProjTestDrift}{n/a}
\providecommand{\OracleProjHeadNorm}{n/a}
\providecommand{\OracleRemovedHeadEnergy}{n/a}
\providecommand{\TruncBasisRank}{n/a}
\providecommand{\TiedFixedTauFailurePct}{n/a}
\providecommand{\OracleFixedTauFailurePct}{n/a}
\providecommand{\TruncFixedTauFailurePct}{n/a}
\providecommand{\MatchedSeeds}{n/a}
\providecommand{\MatchedFibers}{n/a}
\providecommand{\MatchedRho}{n/a}
\providecommand{\StableChi}{n/a}
\providecommand{\IllChi}{n/a}
\providecommand{\IntrinsicChi}{n/a}
\providecommand{\StableInvariantNorm}{n/a}
\providecommand{\IllInvariantNorm}{n/a}
\providecommand{\StableDeployedDrift}{n/a}
\providecommand{\StableSelectedDrift}{n/a}
\providecommand{\IllDeployedDrift}{n/a}
\providecommand{\IllSelectedDrift}{n/a}
\providecommand{\IntrinsicDeployedDrift}{n/a}
\providecommand{\IntrinsicSelectedDrift}{n/a}
\providecommand{\StableDriftReduction}{n/a}
\providecommand{\IllDriftReduction}{n/a}
\providecommand{\IntrinsicDriftReduction}{n/a}
\providecommand{\StableSelectedBacc}{n/a}
\providecommand{\IllSelectedBacc}{n/a}
\providecommand{\IntrinsicSelectedBacc}{n/a}
\providecommand{\StableExactBacc}{n/a}
\providecommand{\IllExactBacc}{n/a}

\providecommand{\DeployedScoreMismatch}{n/a}
\providecommand{\SweepGeometries}{n/a}
\providecommand{\IntrinsicSweepSpearman}{n/a}
\providecommand{\ConditioningSweepSpearman}{n/a}
\providecommand{\IntrinsicLowResidualRatio}{n/a}
\providecommand{\IntrinsicHighResidualRatio}{n/a}
\providecommand{\StableSweepResidualRatio}{n/a}
\providecommand{\IllSweepResidualRatio}{n/a}

\providecommand{\LearningMinTrain}{n/a}
\providecommand{\LearningMaxTrain}{n/a}
\providecommand{\StableRatioSixteen}{n/a}
\providecommand{\StableRatioMax}{n/a}
\providecommand{\IllRatioSixteen}{n/a}
\providecommand{\IllRatioMax}{n/a}
\providecommand{\IntrinsicRatioMax}{n/a}
\providecommand{\UntiedSystems}{n/a}
\providecommand{\UntiedTriples}{n/a}
\providecommand{\UntiedMedianResidualSpearman}{n/a}
\providecommand{\UntiedMedianCrossGramSpearman}{n/a}
\providecommand{\UntiedPooledResidualSpearman}{n/a}
\providecommand{\UntiedPooledCrossGramSpearman}{n/a}

\providecommand{\GlobalEqualRatio}{n/a}
\providecommand{\GlobalRandomOptimizedRatio}{n/a}
\providecommand{\GlobalRandomTiedRatio}{n/a}
\providecommand{\GlobalLeverageSpearman}{n/a}

\providecommand{\BudgetStableKappa}{n/a}
\providecommand{\BudgetIllKappa}{n/a}
\providecommand{\BudgetIntrinsicKappa}{n/a}
\providecommand{\ImportanceModels}{n/a}
\providecommand{\ImportanceSteps}{n/a}
\providecommand{\ImportanceKept}{n/a}
\providecommand{\ImportanceTauUniformMin}{n/a}
\providecommand{\ImportanceTauUniformMax}{n/a}
\providecommand{\ImportanceTauNonuniformMax}{n/a}
\providecommand{\ImportanceAlignUniform}{n/a}
\providecommand{\ImportanceAlignNonuniform}{n/a}
\providecommand{\ImportanceRhoRatioUniform}{n/a}
\providecommand{\ImportanceRhoRatioNonuniform}{n/a}
\providecommand{\ImportanceRhoTopNonuniformMin}{n/a}
\providecommand{\ImportanceRsqLevelUniform}{n/a}
\providecommand{\ImportanceRsqLevelPooled}{n/a}
\providecommand{\ImportanceRsqStructurePooled}{n/a}
\providecommand{\ImportanceWelchGapUniform}{n/a}
\providecommand{\ImportanceWelchGapNonuniform}{n/a}
}
\IfFileExists{pilot_macros.tex}{\newcommand{\PilotFibers}{300}
\newcommand{\PilotTrainFibers}{150}
\newcommand{\PilotAuditedLanguages}{4}
\newcommand{\PilotHeldOutLanguages}{3}
\newcommand{\PilotDimension}{384}
\newcommand{\PilotContrastRank}{221}
\newcommand{\PilotKappa}{0.979}
\newcommand{\PilotReaderNorm}{4.9}
\newcommand{\PilotSelectedRidgeRelative}{3.16}
\newcommand{\PilotSelectedReaderNorm}{1.07}
\newcommand{\PilotAuditedDeployedDrift}{0.122}
\newcommand{\PilotAuditedSelectedDrift}{0.114}
\newcommand{\PilotHeldOutDeployedDrift}{0.213}
\newcommand{\PilotHeldOutSelectedDrift}{0.186}
\newcommand{\PilotAuditedReductionPct}{7}
\newcommand{\PilotHeldOutReductionPct}{13}
\newcommand{\PilotProjectedReductionPct}{-47}
\newcommand{\PilotProjectedUtility}{0.26}
\newcommand{\PilotSelectedUtility}{0.92}
}{\providecommand{\PilotFibers}{300}
\providecommand{\PilotTrainFibers}{150}
\providecommand{\PilotAuditedLanguages}{4}
\providecommand{\PilotHeldOutLanguages}{3}
\providecommand{\PilotDimension}{384}
\providecommand{\PilotContrastRank}{221}
\providecommand{\PilotKappa}{0.979}
\providecommand{\PilotReaderNorm}{4.9}
\providecommand{\PilotSelectedRidgeRelative}{3.16}
\providecommand{\PilotSelectedReaderNorm}{1.07}
\providecommand{\PilotAuditedDeployedDrift}{0.122}
\providecommand{\PilotAuditedSelectedDrift}{0.114}
\providecommand{\PilotHeldOutDeployedDrift}{0.213}
\providecommand{\PilotHeldOutSelectedDrift}{0.186}
\providecommand{\PilotAuditedReductionPct}{7}
\providecommand{\PilotHeldOutReductionPct}{13}
\providecommand{\PilotProjectedReductionPct}{-47}
\providecommand{\PilotProjectedUtility}{0.26}
\providecommand{\PilotSelectedUtility}{0.92}
}

\hypersetup{
    colorlinks=true,
    linkcolor=blue,
    citecolor=blue,
    urlcolor=blue,
    pdftitle={Semantic Fibers and Cross-Gram Interference:
A Calculus of Safety Drift in Overcomplete Representations},
    pdfauthor={Mohammed AHNOUCH and Lotfi Elaachak}
}

\title{Semantic Fibers and Cross-Gram Interference:\\
A Calculus of Safety Drift in Overcomplete Representations}

\author{
    Mohammed AHNOUCH
    \\
    Université Paris 1
    \\
    Paris, France
    \and
    Lotfi Elaachak
    \\
    Faculty of Science and Technology of Tangier
    \\
    Abdelmalek Essaadi University
    \\
    Tangier, Morocco
}

\date{}

\begin{document}

\maketitle
\begin{abstract}
A deployed language model can refuse a harmful request in English and comply
with its faithful translation. Fix the audited equivalence relation, the layer,
the feature dictionary and the scoring head, and that drift becomes exact
linear algebra: it is a cross-Gram functional of the within-fiber contrast, its
worst admissible value is a support function, and invariance of the margin is
an annihilator condition. What the paper adds to the measurement is the verdict
that follows it. An intrinsic calibrated exposure $\chi$, governed by the
leverage duality $\chi^2=1/\ell-1$, sorts an observed drift into three cases
that call for different work: a reader fault that recalibration removes, an
exact fix too ill-conditioned to trust, and a representation-level collision no
readout removes. Matched observed exposure can hide opposite verdicts, and the
same rank condition covers cone-valued safety heads. An untied order-swap
identity turns interventions into a test of the linear control interface, and
its residual, estimated on calibration states, predicts a distinct
three-control composition error on new states and targets (median Spearman
$\UntiedMedianResidualSpearman$ against $\UntiedMedianCrossGramSpearman$ for
the static cross-Gram baseline). Two closed-form repairs act on the head alone,
and two limits bound what any such repair can reach: a column permutation moves
block exposure while fixing the entire Gram spectrum, and reader-optimized
cross-talk obeys the sharp rank floor $n(n-r)/r$, attained by equal-leverage
frames. In trained frames the level of superposition predicts vulnerability
only while feature importance is uniform, the regime that holds frames on the
tight-frame floor. Deterministic synthetic audits verify the identities out of
sample, with intrinsic exposure predicting the repair floor (Spearman
$\IntrinsicSweepSpearman$) and conditioning predicting usable repair
($\ConditioningSweepSpearman$); on one real multilingual encoder with audited
translation fibers the diagnosis returns the conditioning regime, where the
exact invariant reader is harmful out of sample and a validation-selected
regularized reader is not. Each quantity is computed for a declared quotient,
representation, metric, dictionary, head, threshold and contrast model.

\end{abstract}

\section{Introduction}
\label{sec:intro}

A deployed language model can refuse a harmful request in English and comply
with the same request in translation. Translating unsafe prompts into
low-resource languages elicits harmful completions from GPT-4
\citep{yong2024lowresource}, and black-box multilingual red-teaming measures
the same uneven safety-alignment coverage across languages on deployed and
open models \citep{wang2024languages,deng2024multilingual}. The symptom has a
linear substrate: refusal behavior in English prompts is mediated by a
low-dimensional direction that is causally necessary and sufficient
\citep{arditi2024refusal}, and such directions transfer across safety-aligned
languages \citep{wang2025refusal}. The gap between the substrate and the
symptom has since been measured directly: on audited translation pairs the
refusal signal retains a small fraction of its English strength in
low-resource languages while the prompts themselves stay semantically aligned,
so the concept is present but is not routed to the safety mechanism
\citep{oppong2026illusion}. Transporting an English refusal direction across
languages raises refusal rates without retraining \citep{stein2026babelsteering},
which is a repair of exactly the kind this paper prices.

Behavior and direction transfer leave the diagnosis open, and this paper takes
it up: given an
audited equivalence relation and a fixed local representation, which geometric
quantity governs the drift of the refusal margin, and is the observed exposure
removable by changing only the reader? The multilingual jailbreak becomes an
auditable linear-algebra problem: fixing the equivalence relation, the layer,
the dictionary, and the deployed head turns same-meaning drift into an exactly
computable object with a certificate, a three-way diagnosis, an interventional
test, two repairs, and a sharp limit.

A refusal rule should answer to safety-relevant intent, not to the language in
which that intent is expressed. Let $X$ be a prompt set, let $x\sim x'$ mean that
$x$ and $x'$ have the same \emph{externally audited} safety-relevant intent, and
let $\pi:X\to X/{\sim}$ be the quotient map. The relation $\sim$ is fixed before
the model is inspected: the model supplies neither the equivalence classes nor
the evidence of its own invariance on them.

Throughout, refusal means $M(x)>0$. We distinguish
\begin{align*}
\text{margin invariance:}\quad & M(x)=M(x')
  && \text{for all } x\sim x',\\
\text{decision invariance:}\quad & \mathbf 1\{M(x)>0\}=\mathbf 1\{M(x')>0\}
  && \text{for all } x\sim x',\\
\text{directed unsafe failure:}\quad & M(x_{\mathrm{en}})>0\ge M(x_\ell)
  && \text{on an audited unsafe fiber.}
\end{align*}
The annihilator criterion below characterizes margin invariance. Margin
invariance implies decision invariance; the converse fails, because a nonconstant
margin can stay on one side of the threshold throughout a fiber. A multilingual
jailbreak is the directed unsafe failure event, a sign change along one fiber,
not a large excursion in representation space.

A large within-fiber drift has three mathematically different explanations. The
deployed reader may be unnecessarily aligned with nuisance directions even
though another calibrated reader would separate them. The representation may
instead place the safety signal inside the nuisance span, so that no linear
reader can preserve the safety response while annihilating the nuisance block.
A third regime lies between these two: exact separation exists, but only
through a high-norm reader that amplifies reconstruction error and hidden-state
noise. These regimes call for different engineering work, and the quantity that tells
them apart is the intrinsic exposure, since a tied Gram row and a behavioral
disagreement rate read the same in all three. The practical
message can be stated before any frame terminology:
\begin{quote}
\emph{High current exposure with low intrinsic exposure is a reader problem;
high intrinsic exposure is a representation problem; low intrinsic exposure
with a large invariant-reader norm is a conditioning problem.}
\end{quote}

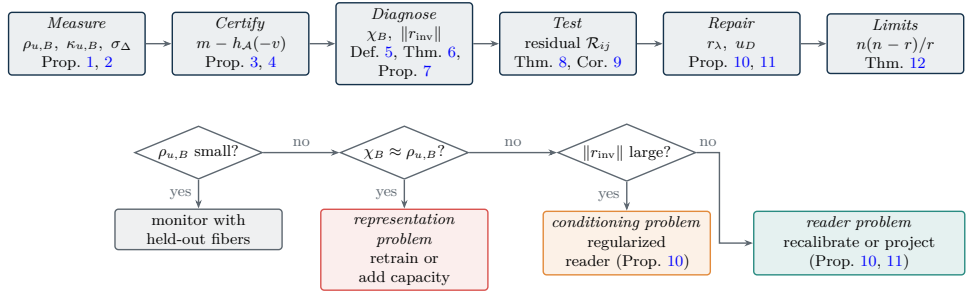
\begin{figure}[t]
\centering
\resizebox{\linewidth}{!}{%
\begin{tikzpicture}[
  font=\footnotesize,
  stage/.style={draw=DeckNavy, thick, rounded corners=2pt, fill=DeckNavy!8,
    minimum height=2.1em, text width=6.6em, align=center, inner sep=3pt},
  verdict/.style={draw, thick, rounded corners=2pt, minimum height=2.0em,
    text width=8.2em, align=center, inner sep=3pt},
  q/.style={draw=DeckGray, thick, diamond, aspect=2.4, inner sep=1pt,
    align=center, fill=white},
  arr/.style={-{Stealth[length=5pt]}, thick, DeckGray},
]
\node[stage] (m) {\emph{Measure}\\ $\rho_{u,B},\ \kappa_{u,B},\ \sigma_\Delta$\\ Prop.~\ref{thm:drift},~\ref{thm:block}};
\node[stage, right=1.35em of m] (c) {\emph{Certify}\\ $m-h_{\calA}(-v)$\\ Prop.~\ref{thm:certificate},~\ref{thm:dictionary}};
\node[stage, right=1.35em of c] (dg) {\emph{Diagnose}\\ $\chi_B,\ \norm{\rinv}$\\ Def.~\ref{def:chi}, Thm.~\ref{thm:leverage}, Prop.~\ref{prop:matched}};
\node[stage, right=1.35em of dg] (t) {\emph{Test}\\ residual $\calR_{ij}$\\ Thm.~\ref{thm:untied}, Cor.~\ref{cor:tied}};
\node[stage, right=1.35em of t] (rp) {\emph{Repair}\\ $\rlam,\ u_D$\\ Prop.~\ref{thm:ridge},~\ref{thm:projection}};
\node[stage, right=1.35em of rp] (l) {\emph{Limits}\\ $n(n-r)/r$\\ Thm.~\ref{thm:global}};
\foreach \a/\b in {m/c, c/dg, dg/t, t/rp, rp/l}{\draw[arr] (\a) -- (\b);}
\node[q, below=2.6em of c, xshift=-2.2em] (q1) {$\rho_{u,B}$ small?};
\node[verdict, draw=DeckGray, fill=DeckGray!10, below=1.5em of q1] (mon) {monitor with held-out fibers};
\node[q, right=4.0em of q1] (q2) {$\chi_B\approx\rho_{u,B}$?};
\node[verdict, draw=DeckRed, fill=DeckRed!10, below=1.5em of q2] (rep) {\emph{representation problem}\\ retrain or add capacity};
\node[q, right=4.6em of q2] (q3) {$\norm{\rinv}$ large?};
\node[verdict, draw=DeckOrange, fill=DeckOrange!12, below=1.5em of q3] (cond) {\emph{conditioning problem}\\ regularized reader (Prop.~\ref{thm:ridge})};
\node[verdict, draw=DeckTeal, fill=DeckTeal!12, text width=10.5em, right=2.2em of cond] (read) {\emph{reader problem}\\ recalibrate or project\\ (Prop.~\ref{thm:ridge},~\ref{thm:projection})};
\draw[arr] (q1) -- node[left]{yes} (mon);
\draw[arr] (q1) -- node[above]{no} (q2);
\draw[arr] (q2) -- node[left]{yes} (rep);
\draw[arr] (q2) -- node[above]{no} (q3);
\draw[arr] (q3) -- node[left]{yes} (cond);
\draw[arr] (q3.east) -- node[above]{no} ++(1.2em,0) |- (read.west);
\end{tikzpicture}%
}
\caption{The audit arc (top) and the three-regime triage it supports (bottom).
Each stage names its governing results and the quantity it contributes; the
triage runs on quantities computable before any repair is attempted.}
\label{fig:taxonomy}
\end{figure}

The paper organizes its results as one workflow, summarized in
\cref{fig:taxonomy}: measure exposure, certify decisions, diagnose
repairability, test the causal interface, repair, and know the limits.
The results form one workflow, drawn in \cref{fig:taxonomy}. Same-fiber drift is
the cross-Gram row applied to the contrast, and constancy of the margin is
exactly an annihilator condition, so exposure is read off the geometry rather
than searched for (\cref{thm:drift,thm:block}); the budget in force selects
which constant governs it, and the feature-coordinate, hidden-state and
stochastic budgets give $\rho_{u,B}$, $\kappa_{u,B}$ and $\sigma_\Delta$. Worst
admissible drift is then a support function, which certifies decisions and keeps
its meaning under an estimated dictionary with one explicit residual
(\cref{thm:certificate,thm:dictionary}). The intrinsic calibrated exposure
$\chi_B$ carries the verdict: it satisfies $\chi_B^2=1/\ell_B-1$ for the
relative leverage $\ell_B$, feasibility separates from stability through the
reader norm $1/\sin\theta_{s,B}$, and two geometries with identical observed
exposure can call for opposite work
(\cref{def:chi,thm:leverage,prop:matched}). An untied order-swap identity
predicts the commutation defect of calibrated read and write interventions
exactly, and the observed-minus-predicted residual reports how far a steering
stack follows the linear interface that every readout repair assumes
(\cref{thm:untied,cor:tied}). Two closed-form repairs act on the head alone, a
regularized calibrated reader tracing the frontier between the deployed readout
and exact cancellation, and a least-change projection carrying a held-out
transfer bound (\cref{thm:ridge,thm:projection}). Two limits frame both: a
column permutation moves block exposure while holding the entire Gram spectrum
fixed, and reader-optimized cross-talk obeys the sharp rank floor $n(n-r)/r$,
attained by equal-leverage frames (\cref{thm:global}); in trained frames
(\cref{sec:importance}) the level of superposition identifies exposure while
feature importance is uniform. Each stage closes with measurements built around
out-of-sample consequences of its identities, and \cref{sec:recipe} turns the
same geometry into an audit that runs on a real stack.
Each stage closes with a numerical audit designed around out-of-sample
consequences of its identities rather than confirmations of them, and an
engineering recipe (\cref{sec:recipe}) turns the same geometry into action.

A running case makes the arc concrete. A provider's guard score refuses a
harmful request in English and passes its faithful translation, and the
response team holds paired activations at the scored layer, the score
direction, and optionally a sparse-autoencoder dictionary. Current exposure is
then a short computation on their own tensors, from the dictionary or from
paired hidden differences. The intrinsic exposure $\chi$ answers the first
strategic question, whether any calibrated head-side fix exists at that layer,
before a single reader is trained; the invariant-reader norm prices whether
the exact fix is usable or only nominal, and the order-swap residual checks
that their steering stack obeys the linear interface every such repair
assumes. Low $\chi$ with moderate conditioning routes the incident to the
regularized reader or the least-change projection with its held-out bound;
high $\chi$ ends the search for a better head and moves the repair budget to
the representation. The calculus and its out-of-sample validation are
established on synthetic geometry, and carrying them onto a team's model needs
their audited fiber set at the layer, dictionary, head and equivalence relation
they declare.

\section{Fibers, writers, and readers}
\label{sec:setup}

Let $\calS=X/{\sim}$ and $F_s=\pi^{-1}(s)$ the fiber over $s$. Concretely a fiber
collects a prompt with its audited translations and paraphrases; the
equivalence relation is an external scientific and governance choice, defining
which changes the policy declares irrelevant. For a base point $x_0\in F_s$ let
the \emph{realized} contrast set and its \emph{linearized} span be
\begin{equation}
\calA_s(x_0)=\{z(x)-z(x_0):x\in F_s\},\qquad
\calT_s=\spanop\calA_s(x_0)=\calC_s,
\label{eq:contrast}
\end{equation}
where $\calC_s=\spanop\{z(x)-z(x'):x,x'\in F_s\}$; the two symbols separate the
realized set from its span. Radius statements are stated for Euclidean balls in
$\calT_s$: they are exact for the linearized local model and become prompt-level
certificates only when the contrast is realized or an audit supplies the feasible
set. A ball in $\calT_s$ may contain directions no prompt realizes.

For a block $B\subset\{1,\dots,n\}$ of language coordinates, $P_B$ is the
coordinate projection, $W_B$ the columns of $W$ indexed by $B$, and
$\calL_{s,B}=\calC_s\cap\R^B$ the admissible block subspace (an intersection of
subspaces, hence a subspace). If $u=W\alpha$ then, whenever $W\alpha=W\alpha'$,
$G\alpha=G\alpha'$ exactly, since $G(\alpha-\alpha')=W^\top W(\alpha-\alpha')
=W^\top(W\alpha-W\alpha')=0$; the block quantities below are therefore
well-defined in $\alpha$.

\label{sec:writers-readers}
At a fixed layer, on the prompt family under study, the local model is
\begin{equation}
h(x)=Wz(x)+\varepsilon(x),\qquad
W=[w_1,\dots,w_n]\in\R^{d\times n},\qquad
G=W^\top W,
\label{eq:frame}
\end{equation}
with unit columns unless stated otherwise and $\norm{\varepsilon(x)}\le\eta$.
The columns of $W$ are \emph{writers}: directions along which feature activity
changes the hidden state. A reader matrix $R=[r_1,\dots,r_n]\in\R^{d\times n}$
defines measurements $r_i^\top h$; readers and writers need not coincide, as is
typical for sparse-autoencoder encoders and decoders
\citep{cunningham2023sae,bricken2023monosemanticity}, linear probes and
activation interventions, or the oblique analysis/synthesis systems of
classical frame theory
\citep{christensen2016frames,eldar2003oblique,eldar2006oblique}. The
self-calibration convention $r_i^\top w_i=1$ fixes the intended response of
each reader to its own writer, and the \emph{cross-Gram} matrix
\begin{equation}
C=R^\top W,\qquad C_{ij}=r_i^\top w_j,
\label{eq:crossgram}
\end{equation}
measures directional read/write cross-talk; the tied case $R=W$ with unit
columns gives the ordinary Gram matrix $C=G$.

The deployed safety head is a reader $u$ with margin $M(x)=u^\top h(x)-\tau$,
and a designated safety writer $w_s$ anchors the audited signal. When
$u^\top w_s=1$, which includes the tied head $u=w_s$ with unit columns, the
deployed head competes with every calibrated reader, and the intrinsic exposure
of \cref{def:chi} satisfies $\chi_B\le\rho_{u,B}$. Write $v=W^\top u$ for the
pullback of the head and $v_B=W_B^\top u=P_Bv$ for its block row; if $u$ is the
$s$-th reader column, $v_B$ is the cross-Gram row segment $C_{sB}$.

\begin{figure}[t]
\centering
\resizebox{0.62\linewidth}{!}{%
\begin{tikzpicture}[
  font=\scriptsize,
  layer/.style={draw=DeckGray, thick, rounded corners=2pt, fill=DeckGray!8,
    minimum width=4.4em, minimum height=1.7em},
  arr/.style={-{Stealth[length=4pt]}, thick, DeckGray},
]
\node[layer] (l1) {layer $\ell-1$};
\node[layer, right=2.1em of l1] (l2) {layer $\ell$};
\node[layer, right=2.1em of l2] (l3) {layer $\ell+1$};
\draw[arr] (l1) -- (l2); \draw[arr] (l2) -- (l3);
\node[below=0.9em of l1, align=center, text width=8em] (p) {same audited intent,\\ two languages};
\draw[arr, DeckBlue] (p.north) -- (l1.south);
\coordinate (z) at ($(l2.north)+(-0.6em,3.1em)$);
\draw[DeckNavy, thick] (z) circle (1.9em);
\draw[DeckNavy, thick] ($(z)+(-1.05em,-1.6em)$) -- ($(l2.north)+(-0.25em,0)$);
\fill[DeckBlue] ($(z)+(-0.45em,0.25em)$) circle (1.6pt) node[above left=-2pt]{$h(x_{\mathrm{en}})$};
\fill[DeckBlue] ($(z)+(0.55em,-0.15em)$) circle (1.6pt) node[below right=-2pt]{$h(x_{\mathrm{fr}})$};
\coordinate (o) at ($(z)+(6.1em,0.2em)$);
\foreach \a in {40,62,82}{\draw[arr, DeckGray!70] (o) -- ++(\a:2.4em);}
\draw[arr, DeckTeal, very thick] (o) -- ++(107:2.5em);
\draw[arr, DeckTeal, very thick] (o) -- ++(130:2.3em);
\node[DeckTeal] at ($(o)+(118:3.3em)$) {$W_B$};
\draw[arr, DeckRed, very thick] (o) -- ++(8:2.7em) node[right]{$w_s$};
\draw[arr, DeckOrange, very thick] (o) -- ++(-24:2.7em) node[right]{reader $u$};
\draw[decorate, decoration={brace, amplitude=3pt}, DeckTeal]
  ($(o)+(130:2.6em)$) -- ($(o)+(105:2.9em)$);
\node[align=left, text width=9.6em, anchor=west] (setter)
  at ($(o)+(4.9em,1.6em)$)
  {$T_i^a$: read along $r_i$ (dashed), write along $w_i$ (solid)};
\draw[arr, DeckPurple, dashed] (setter.south west) to[bend left=18] ($(o)+(0.3em,0.55em)$);
\draw[arr, DeckPurple] ($(o)+(0.7em,0.1em)$) to[bend right=18] ($(setter.west)+(0,-0.9em)$);
\end{tikzpicture}%
}
\caption{The framework in a neural network. A semantic fiber places two
same-intent prompts at nearby hidden states in the residual stream; writers
move states, readers measure them, and an untied intervention reads along
$r_i$ but writes along $w_i$. The drift of the safety margin between the two
states is the cross-Gram functional of \cref{thm:drift}.}
\label{fig:network}
\end{figure}
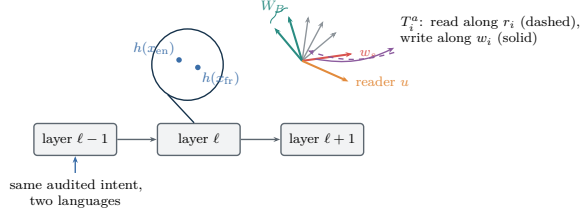

\label{sec:network-view}
In machine-learning terms, $h(x)$ is a fixed-layer activation, $W$ a learned
feature dictionary in the reading supplied by superposition and
sparse-autoencoder work \citep{elhage2022toy,templeton2024scaling}, for
instance a sparse-autoencoder decoder, $z(x)$ its code, and $u$ a linear refusal probe with threshold $\tau$; the results compute
worst-case and average-case refusal-margin drift under same-meaning language
changes, in the sense in which a certified radius bounds loss change under a
bounded perturbation. \Cref{fig:network} draws this view, and the paragraph
below maps each object to its usual implementation. Column
normalization fixes one scaling convention, and every radius below is relative
to the declared metric in that fixed coordinate system; comparisons across
encoders must report the normalization and metric used for contrasts.

Each object in this framework names something an engineer already has. A
semantic fiber is an audited set of translations, paraphrases, modality variants
or tool-schema aliases, that is, inputs a policy says should receive the same
treatment. A writer $w_j$ is a decoder column of a sparse autoencoder, an
activation direction or an intervention direction, and it says how feature $j$
moves the hidden state; a reader $r_i$ is a probe, a classifier head, an encoder
direction or the score direction itself, and it says how feature $i$ is
measured. Their product, the cross-Gram $C=R^\top W$, is what a reader records
when a writer is driven, so it is directional read and write cross-talk. Current
exposure $\rho_{u,B}=\norm{W_B^\top u}$ is how much the deployed reader listens
to the nuisance block, while intrinsic exposure $\chi_B$, the least exposure
available under the calibration $r^\top w_s=1$, is whether a head-side repair can
remove that response at all. Conditioning $\norm{\rinv}$, the norm of the exact
invariant reader, prices how much residual noise and estimation error such a
repair amplifies. The order-swap residual $\calR_{ij}$, an observed order
difference minus its linear cross-Gram prediction, reports whether controls
compose the way the linear model says they do.

The experiments are deterministic synthetic mechanism checks, operating
entirely on dictionaries, labels, and heads fixed by construction, independent
of prompt text, jailbreak search, or any deployed model. A language is an
abstract block of feature coordinates whose variation fixes the audited safety
coordinate. Ten pipeline stages run in fresh processes from local
pseudorandom generators in CPU float64. The matched benchmark uses
$\MatchedSeeds$ rotation seeds and $\MatchedFibers$ fibers per geometry with
train/validation/test splits of $4000/2000/2000$ by fiber; the block experiment
uses $d=\DimD$, $n=\NumFeatures$, $|B|=\LangBlockSize$, level
$n/d=\BlockLevel$, and eight frame seeds, with every sampled contrast rescaled
so that $\norm{W_Bc}_2=2.45$, fixing hidden distance by construction; the drift
experiment uses $\NumFibers$ fibers, fixed threshold $\tau=0.65$, and noise
$\eta=\NoiseEta$; the repair experiments select every penalty on validation
under a prespecified utility constraint and evaluate once on the untouched
test split. Symmetric $\pm1$ labels make zero the population threshold for
every calibrated reader in the matched benchmark, so no operating-point degree
of freedom is fitted there. Replication units are the rotation seed (matched,
sweep, learning), the frame seed (block, repair), the feature pair (tied
commutator), and the reader-writer system (untied). The accompanying code
regenerates every number. The audit is designed
around out-of-sample consequences: readers are estimated on training fibers,
selected on validation fibers, and evaluated once on untouched test fibers,
and the untied diagnostic predicts a task it was not fitted on.

\section{Exposure and certificates}
\label{sec:measure-certify}

Classical adversarial work first found optimization-based perturbations that
cross decision boundaries and transfer across independently trained models
\citep{szegedy2014intriguing}, an effect later attributed to near-linear
behavior in high dimensions and distilled into a closed-form attack
\citep{goodfellow2015adversarial}. Textual attacks construct concrete
adversarial strings by gradient-guided discrete search, neither
meaning-preserving by construction: single-input character or word edits
\citep{ebrahimi2018hotflip} and dataset-universal trigger tokens
\citep{wallace2019triggers}. This section computes the local obstruction
directly once the fiber is fixed, rather than searching for attacks: drift is
a linear functional, invariance is an annihilator condition, and the worst
admissible drift is a support function.

\subsection{Drift and sensitivity constants}
\label{sec:measure}

The first result is exact linear algebra: it identifies same-fiber drift and the
condition for margin invariance.

\begin{proposition}[Same-fiber drift and the annihilator]
\label{thm:drift}
Fix a fiber $F_s$ and let $v=W^\top u$.
\begin{enumerate}[label=\textup{(\alph*)},leftmargin=*,itemsep=1pt,topsep=2pt]
\item If $\varepsilon=0$ on $F_s$, then $M(x)-M(x')=v^\top(z(x)-z(x'))$ for all
$x,x'\in F_s$; if $u=W\alpha$ this equals $\alpha^\top G(z(x)-z(x'))$, and if
$u$ is the $s$-th column of a reader system $R$ and the contrast is supported
on the block $B$, it equals the cross-Gram row form $C_{sB}\,c$.
\item The margin is constant on $F_s$ if and only if $v\in\calC_s^\perp$,
equivalently $P_{\calC_s}v=0$.
\item If $\norm{\varepsilon(x)}\le\eta$ on $F_s$, then
$\bigl|M(x)-M(x')-v^\top(z(x)-z(x'))\bigr|\le 2\eta\norm{u}$.
\end{enumerate}
\end{proposition}

\begin{claimproof}
\item $M(x)-M(x')=u^\top W(z(x)-z(x'))=v^\top(z(x)-z(x'))$ when $\varepsilon=0$.
\by{$M=u^\top Wz-\tau$}
\item $u=W\alpha\Rightarrow v=W^\top W\alpha=G\alpha$. \by{definition of $G$}
\item For $u=r_s$ and a block contrast $c$, $v^\top(z(x)-z(x'))
=(W_B^\top r_s)^\top c=C_{sB}\,c$. \by{definition of $C$}
\item The drift vanishes on all pairs iff $v$ annihilates $\calC_s$, i.e.
$v\in\calC_s^\perp$. \by{$\calC_s$ spans the contrasts}
\item With residuals, $M(x)-M(x')=v^\top(z(x)-z(x'))+u^\top(\varepsilon(x)-\varepsilon(x'))$
and $|u^\top(\varepsilon(x)-\varepsilon(x'))|\le 2\eta\norm{u}$. \by{Cauchy-Schwarz}
\end{claimproof}

Invariance thus asks the head to annihilate within-fiber contrasts, a
substantially weaker requirement than collapse of the hidden states themselves.
The exposed direction is $v_s=P_{\calC_s}v$. The next result gives the
worst-case constants: the threat metric determines which one governs a language
block, and a stochastic contrast model contributes a third, distinct scale.

\begin{proposition}[Metric-specific block sensitivity]
\label{thm:block}
Assume $\varepsilon=0$ and write $v_B=W_B^\top u$.
\begin{enumerate}[label=\textup{(\alph*)},leftmargin=*,itemsep=1pt,topsep=2pt]
\item For a feature-coordinate Euclidean budget,
\begin{equation}
\sup_{\norm{c}_2\le r}|u^\top W_Bc|=r\,\rho_{u,B},\qquad
\rho_{u,B}=\norm{v_B}_2\ \ (=\norm{P_BG\alpha}\ \text{if }u=W\alpha),
\label{eq:rho}
\end{equation}
and more generally, for a positive-semidefinite budget $c^\top Qc\le r^2$ the
supremum is finite iff $v_B\in\range(Q)$, in which case it equals
$r\sqrt{v_B^\top Q^\dagger v_B}$. When $u=w_{\isafe}$ is tied to a single
feature, $\rho_{\isafe,B}=(\sum_{j\in B}G_{j\isafe}^2)^{1/2}$, the block row
norm.
\item For a hidden-state budget, with $G_{BB}=W_B^\top W_B$ and $\dagger$ the
Moore-Penrose inverse,
\begin{equation}
\sup_{\norm{W_Bc}_2\le r}|u^\top W_Bc|=r\,\kappa_{u,B},\qquad
\kappa_{u,B}=\sqrt{v_B^\top G_{BB}^\dagger v_B}=\norm{P_{\col(W_B)}u}_2.
\label{eq:kappa}
\end{equation}
\item For a random block contrast $c$ with mean $\mu_B$ and covariance
$\Sigma_B$: $\E[\Delta M]=v_B^\top\mu_B$ and
$\Var(\Delta M)=v_B^\top\Sigma_Bv_B=:\sigma_\Delta^2$; if $\mu_B=0$ and
$\Sigma_B=\sigma_c^2\Id$ with unit columns then
$\Var(\Delta M)=\sigma_c^2\rho_{u,B}^2$ while $\E\norm{W_Bc}_2^2=\sigma_c^2|B|$.
Under the symmetric construction $M_{\mathrm{en}}=m_0-\tfrac12\Delta M$,
$M_{\mathrm{other}}=m_0+\tfrac12\Delta M$, the directed event
$M_{\mathrm{en}}>0\ge M_{\mathrm{other}}$ is $\Delta M\le-2m_0$; if the
conditional drift is Gaussian with standard deviation $\sigma_\Delta$ then
$\Prob(M_{\mathrm{en}}>0\ge M_{\mathrm{other}}\mid m_0)=\Phi(-2m_0/\sigma_\Delta)$.
\end{enumerate}
\end{proposition}

\begin{claimproof}
\item $u^\top W_Bc=v_B^\top c$, so \eqref{eq:rho} is Euclidean duality with
equality at $c=r\,v_B/\norm{v_B}$. \by{Cauchy-Schwarz}
\item For $c^\top Qc\le r^2$, split $c$ along $\range(Q)$ and
$\ker Q$; a component of $v_B$ in $\ker Q$ makes the objective unbounded.
\by{zero-cost direction}
\item Else $v_B=Q^{1/2}q$ with $q=Q^{\dagger/2}v_B$, and
$|v_B^\top c|=|q^\top Q^{1/2}c|\le\norm{q}\sqrt{c^\top Qc}=r\sqrt{v_B^\top Q^\dagger v_B}$.
\by{Cauchy-Schwarz}
\item $Q=G_{BB}$ is admissible since
$v_B=W_B^\top u\in\range(W_B^\top)=\range(G_{BB})$,
and $W_BG_{BB}^\dagger W_B^\top=P_{\col(W_B)}$. \by{\eqref{eq:kappa}}
\item $\Delta M=v_B^\top c$, so $\E[\Delta M]=v_B^\top\mu_B$ and
$\Var(\Delta M)=v_B^\top\Sigma_Bv_B$. \by{linear form}
\item $\Sigma_B=\sigma_c^2\Id\Rightarrow\Var(\Delta M)=\sigma_c^2\norm{v_B}^2
=\sigma_c^2\rho_{u,B}^2$; and $\E\norm{W_Bc}^2=\sigma_c^2\tr G_{BB}=\sigma_c^2|B|$.
\by{$\tr G_{BB}=|B|$}
\item The directed event is $M_{\mathrm{en}}>0\ge M_{\mathrm{other}}$, i.e.
$\Delta M\le-2m_0$, with probability $\Phi(-2m_0/\sigma_\Delta)$.
\by{Gaussian $\Delta M$}
\end{claimproof}

The three constants answer different questions: $\rho_{u,B}$ is worst case per
unit feature-code norm, $\kappa_{u,B}$ per unit hidden-state distance, and
$\sigma_\Delta$ is the stochastic drift scale. They are not interchangeable,
and decision failure additionally depends on the baseline-margin distribution.
The numerical audit separates all three at a fixed overcompleteness level
(\cref{fig:block}), validates the Gaussian failure prediction of part (c), and
traces the sign geometry of the directed failure event (\cref{fig:sign}).

\subsection{Decision certificates and measured exposure}
\label{sec:certify}

The certificate below turns measurement into a decision guarantee against an
audited threat set through its support function, stated as an infimum so that
the identity holds whether or not the bound is attained; attainment itself is
addressed separately when $\calA$ is compact.

\begin{proposition}[Admissible-contrast certificate]
\label{thm:certificate}
Let $\calA\subseteq\calT_s$ be nonempty with support function
$h_\calA(q)=\sup_{c\in\calA}q^\top c$, let $v=W^\top u$, assume $\varepsilon=0$,
and let the unperturbed margin be $m>0$. Writing $\Mt(z)=u^\top Wz-\tau$ for the
feature-space margin,
\begin{equation}
\inf_{c\in\calA}\Mt(z+c)=m-h_\calA(-v).
\label{eq:support-inf}
\end{equation}
Hence: if $h_\calA(-v)<m$ every admissible margin is positive; if $h_\calA(-v)>m$
some admissible perturbation makes it negative; and if $\calA$ is compact a
nonpositive margin is attainable iff $h_\calA(-v)\ge m$, equality being first
contact with the boundary. If $\calA=rK$ with $K$ compact and $h_K(-v)>0$, the
least scale reaching the boundary is $r_*=m/h_K(-v)$; if $h_K(-v)=0$ then
$r_*=+\infty$.
\end{proposition}

\begin{claimproof}
\item $\Mt(z+c)=m+v^\top c$ for every $c$. \by{$\Mt$ affine, $\Mt(z)=m$}
\item $\inf_{c\in\calA}(m+v^\top c)=m-\sup_{c\in\calA}(-v)^\top c=m-h_\calA(-v)$.
\by{$\inf$/$\sup$ duality}
\item The three cases read off the sign of $m-h_\calA(-v)$. \by{\eqref{eq:support-inf}}
\item If $\calA$ compact the supremum is attained, so $h_\calA(-v)=m$ gives a
$c$ with $\Mt(z+c)=0$. \by{Weierstrass}
\item $h_{rK}(-v)=r\,h_K(-v)$, so $m-r h_K(-v)=0$ at $r_*=m/h_K(-v)$.
\by{positive homogeneity}
\end{claimproof}

The Euclidean ball $\calA=\{c\in\calT_s:\norm{c}\le r\}$ is the special case
$h_\calA(-v)=r\norm{v_s}$, giving the linearized feature-space radius
$r_*(x_0)=M(x_0)/\norm{v_s}$; $\Mt$ is a synthetic local extension of the
prompt-defined margin, and this radius changes under nonorthogonal
reparameterizations of code space. If the residual bound of
\cref{thm:drift}(c) holds uniformly on the tube
$\{z(x_0)+c:\norm{c}\le r\}$, no sign flip is certifiable while
$M(x_0)>r\norm{v_s}+2\eta\norm{u}$; without the uniform tube hypothesis the
residual statement is pairwise only.

Certification survives an estimated dictionary with an explicit residual term;
this is what a real learned-dictionary study, working with sparse-autoencoder
estimates of $W$ \citep{cunningham2023sae,templeton2024scaling}, must control.

\begin{proposition}[Approximate dictionary pullback]
\label{thm:dictionary}
Suppose $h(x)=\widehat W\widehat z(x)+\widehat r(x)$ with
$\norm{\widehat r(x)}\le\widehat\eta$ at $x_0,x_1$, and let
$\Delta\widehat z=\widehat z(x_1)-\widehat z(x_0)$. Then
\begin{equation}
\bigl|M(x_1)-M(x_0)-(\widehat W^\top u)^\top\Delta\widehat z\bigr|\le 2\widehat\eta\norm{u}.
\label{eq:pullback}
\end{equation}
For any coefficient $\widehat\alpha$ with $q=u-\widehat W\widehat\alpha$ and
$\widehat G=\widehat W^\top\widehat W$,
$\bigl|M(x_1)-M(x_0)-\widehat\alpha^\top\widehat G\Delta\widehat z\bigr|
\le 2\widehat\eta\norm{u}+|q^\top\widehat W\Delta\widehat z|$; if
$\widehat\alpha=\widehat W^+u$ then $q\perp\col(\widehat W)$ and the second term
vanishes.
\end{proposition}

\begin{claimproof}
\item $M(x_1)-M(x_0)=u^\top\widehat W\Delta\widehat z
+u^\top(\widehat r_1-\widehat r_0)$. \by{threshold cancels}
\item $|u^\top(\widehat r_1-\widehat r_0)|\le 2\widehat\eta\norm{u}$, giving
\eqref{eq:pullback}. \by{Cauchy-Schwarz}
\item $u=\widehat W\widehat\alpha+q$ gives the coefficient form; the
least-squares residual $q=u-\widehat W\widehat W^+u$ is orthogonal to
$\col(\widehat W)$. \by{normal equations}
\end{claimproof}

The direct observable is $\widehat W^\top u$; a real audit reports reconstruction
residuals on both pair members, the encoded-contrast metric and its stability,
and empirical contrast-subspace error, adding a coefficient term only if a
Gram-coordinate representation replaces the exact pullback.

\begin{figure}[tbp]
\centering
\includegraphics[width=0.98\linewidth]{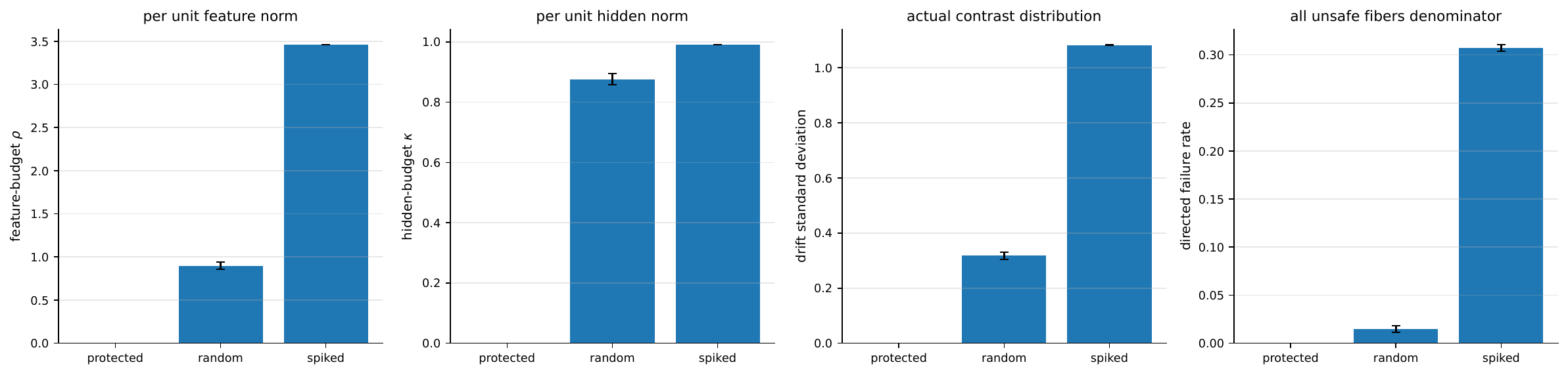}
\caption{Constant-level block families ($n/d=\BlockLevel$, hidden distance fixed
by construction). Feature-budget $\rho$ separates the families far more than the
matching hidden-budget $\kappa$; the directed failure rate tracks the drift
standard deviation, not $\rho$. Error bars are standard errors over eight frame
seeds.}
\label{fig:block}
\end{figure}

Under the protocol of \cref{sec:setup}, \cref{fig:block} shows that, at
the common level $n/d=\BlockLevel$, the
feature-budget norms are
$\ProtectedRhoFeature$, $\RandomRhoFeature$, $\SpikedRhoFeature$ for the
protected, random, and spiked families, a spiked/random ratio of about $3.9$
(squared, about $15$). The random value sits near the isotropic benchmark
$\sqrt{|B|/d}=\RandomIsotropicBenchmark$, and
$\SpikedRhoFeature/\sqrt{|B|}=0.500$ discloses that each spiked block column
carries safety component $\beta=0.5$. But under the fixed-hidden-distance
design the matching constants are the hidden-budget norms
$\ProtectedKappaHidden$, $\RandomKappaHidden$, $\SpikedKappaHidden$
(spiked/random ratio only about $1.13$) and the drift scales
$\ProtectedDriftSd$, $\RandomDriftSd$, $\SpikedDriftSd$. The directed unsafe
rates are $\ProtectedDirectedFailurePct\%$, $\RandomDirectedFailurePct\%$,
$\SpikedDirectedFailurePct\%$. The correct reading is therefore that $n/d$
does not identify block exposure, and that under a hidden-distance budget the
drift distribution, not $\rho$, drives the separation. The same pattern holds
on the matched-exposure families at a fixed hidden radius: the hidden budget
$\kappa=\BudgetStableKappa/\BudgetIllKappa/\BudgetIntrinsicKappa$ climbs to
its ceiling for all three while $\rho$ stays constant by construction, so the
metric choice, not the family, again decides what is being measured. The
compression is structural, not accidental: by \cref{eq:kappa},
$\kappa_{u,B}=\norm{P_{\col(W_B)}u}$ is a projection norm that can never
exceed $\norm{u}=1$, so once a family places the safety writer near the block
span, $\kappa$ saturates, while $\rho$ obeys no such bound. The two constants
live on different scales by identity, not by tuning. This
budget dependence mirrors the arc of the attack literature, from
perturbations found by optimization to the closed-form step licensed by
near-linear behavior \citep{szegedy2014intriguing,goodfellow2015adversarial};
here the local model is linear by declaration, so each declared budget has an
exact worst case, and \cref{thm:certificate} prices the boundary in the same
declared metric.

These drift scales set the standardized boundary distance directly. The mean
unsafe central margin is $\RandomCenterMargin$ (random) and $\SpikedCenterMargin$
(spiked), so the standardized boundary distances $2m_0/\sigma_\Delta$ are
$\RandomStandardizedMargin$ and $\SpikedStandardizedMargin$. A conditional
Gaussian approximation predicts $\RandomGaussianFailurePct\%$ for the random
family, close to the observed $\RandomDirectedFailurePct\%$, and
$\SpikedGaussianFailurePct\%$ for the spiked family; the observed
$\SpikedDirectedFailurePct\%$ then quantifies the non-Gaussian tail. The
undershoot grows with the drift scale, and it is the expected signature of
sparsity: each contrast activates roughly $0.45$ of the block, so the drift is
a sum over a random support whose size varies fiber to fiber, a Gaussian scale
mixture that is leptokurtic and places more mass beyond any fixed boundary
than a single Gaussian at the pooled variance. The fixed-$\sigma_\Delta$
prediction of \cref{thm:block}(c) therefore reads as a floor on the directed
rate, and the sparsity that produces the heavy tail is the same sparsity that
motivates the superposition reading of the dictionary.

\begin{figure}[tbp]
\centering
\includegraphics[width=0.58\linewidth]{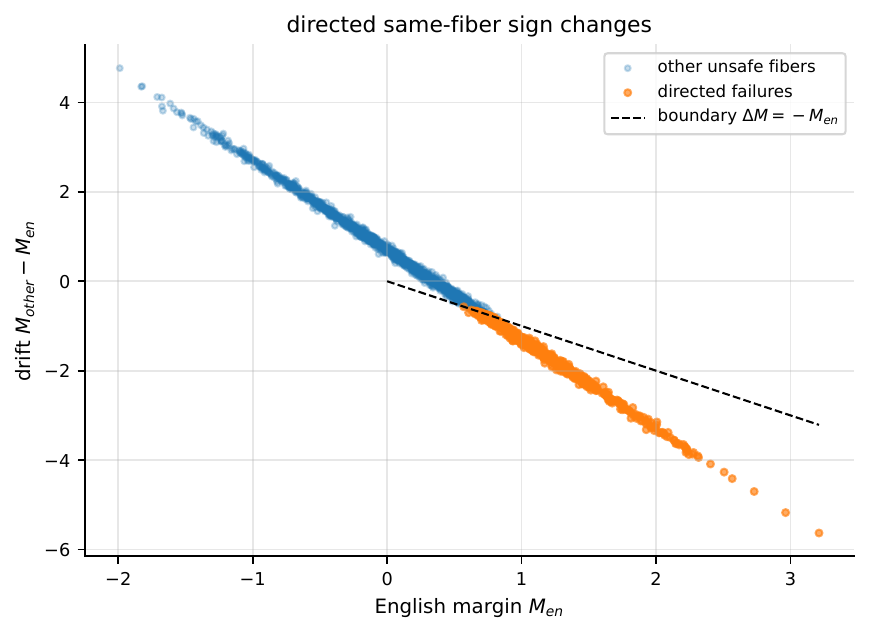}
\caption{Sign geometry of directed failures: the failing fibers are exactly
the points with $\Delta M\le-M_{\mathrm{en}}$, on and below the plotted
boundary; the construction is symmetric, so the reverse directed event has
comparable mass.}
\label{fig:sign}
\end{figure}

\Cref{fig:sign} shows the sign geometry that the exact identity predicts. The
no-noise discrepancy between exact and Gram drift is $\MaxExactGramError$, a
floating-point check. With noise, the mean absolute residual is
$\MeanAbsNoiseResidual$ against the bound $\NoiseResidualBound$, and the $95$th
percentile of $|{\rm residual}|/(2\eta\norm{u})$ is $\ResidualRatioTail$:
typical isotropic residuals concentrate an order of magnitude below the
adversarial-alignment constant, exactly as Cauchy-Schwarz slack should behave,
so the certificate prices the aligned worst case rather than the typical pair.
The
decision-disagreement rate is $\DecisionDisagreementRate\%$; the directed unsafe
population rate is $\DirectedFailureRate\%$, the reverse directed rate
$\ReverseFailureRate\%$ (near-symmetric, as the construction requires), and the
rate conditional on English refusal $\ConditionalDirectedFailureRate\%$; benign
over-refusal is $\BenignOverRefusalEn\%$ in English. The symmetry check is
quantitative: the directed and reverse events are mutually exclusive on each
fiber, and the observed gap between their rates is on the order of one
standard error of its own estimator, precisely the residue finite sampling
should leave of an exact design symmetry; the benign over-refusal rates pass
the same test on the safe side. For any sign disagreement the
AUCs are $\AucOracleDrift$ (oracle absolute drift, gain $0.47$ above chance),
$\AucHiddenDistance$ (hidden distance), $\AucDiagonalProxy$ (diagonal coherence
proxy), and $\AucPermutation$ (permutation control); the first is part of the
exact margin formula and is a calibration check, and the proxy is weak because it
discards signs, cross terms, and the baseline margin. The ladder is the
paper's thesis in four numbers: each rung deletes one layer of cross-Gram
structure, from the full signed functional to magnitude without signs to
diagonal coherence without cross terms to permuted structure, and the
discriminative power decays monotonically with each deletion; the untied
audit of \cref{sec:test} replays the same lesson at the intervention level,
where the signed residual transfers ($\UntiedMedianResidualSpearman$) and the
unsigned static summary does not ($\UntiedMedianCrossGramSpearman$). Within the
$\MarginStrataTotal$ central-margin strata the top-versus-bottom drift ordering
holds in $\GramStrataOrdered$ of $\MarginStrataTotal$ strata, against
$\PermutedStrataOrdered$ of $\MarginStrataTotal$ for the permuted control, which
shows no consistent monotone ordering.

\section{Diagnosis}

Current exposure is a property of the deployed reader. This section asks
whether the representation itself forces that exposure.

\subsection{Intrinsic exposure, leverage duality and conditioning}

\begin{definition}[Intrinsic calibrated exposure and relative leverage]
\label{def:chi}
For a nonzero safety writer $w_s$ and nuisance block $W_B$, define
\begin{equation}
\chi_B:=\min_{r^\top w_s=1}\norm{W_B^\top r}_2.
\label{eq:chi}
\end{equation}
Let $\widetilde W=[w_s,W_B]$, $\widetilde S=\widetilde W\widetilde W^\top$, and
\begin{equation}
\ell_B:=w_s^\top\widetilde S^\dagger w_s,
\label{eq:relativeleverage}
\end{equation}
the safety writer's \emph{relative leverage} with respect to the nuisance
block. Both quantities depend on $w_s$; the subscript $s$ is suppressed.
\end{definition}

\begin{theorem}[Calibrated quotient duality]
\label{thm:leverage}
Let $w_s\ne0$ and let $\theta_{s,B}$ be the angle from $w_s$ to $\spanop(W_B)$.
\begin{enumerate}[label=\textup{(\alph*)},leftmargin=*,itemsep=1pt,topsep=2pt]
\item $\chi_B^2=1/\ell_B-1$, a minimum-norm optimizer is
$\rstar=\widetilde S^\dagger w_s/(w_s^\top\widetilde S^\dagger w_s)$, and every
optimizer is $\rstar+q$ with $q\in\nullsp(\widetilde S)$.
\item $\chi_B=0$ if and only if $w_s\notin\spanop(W_B)$.
\item When $\chi_B=0$, the minimum-norm exactly invariant calibrated reader is
$\rinv=P_{\spanop(W_B)^\perp}w_s/\norm{P_{\spanop(W_B)^\perp}w_s}^2$, and for
unit $w_s$, $\norm{\rinv}=1/\sin\theta_{s,B}$.
\end{enumerate}
\end{theorem}

\begin{claimproof}
\item For calibrated $r$:
$r^\top\widetilde S r=(r^\top w_s)^2+\norm{W_B^\top r}^2=1+\norm{W_B^\top r}^2$.
\by{expand $\widetilde S$}
\item $w_s\in\range(\widetilde S)$, so
$1=(r^\top w_s)^2\le(r^\top\widetilde S r)(w_s^\top\widetilde S^\dagger w_s)
=(1+\norm{W_B^\top r}^2)\,\ell_B$. \by{Cauchy-Schwarz in the $\widetilde S$ seminorm}
\item Rearranging, $\norm{W_B^\top r}^2\ge 1/\ell_B-1$ for every calibrated $r$.
\by{step (2)}
\item $\rstar$ is calibrated and attains equality, since
$\rstar{}^\top\widetilde S\rstar=1/\ell_B$. \by{substitute \eqref{eq:relativeleverage}}
\item Adding $q\in\nullsp(\widetilde S)$ changes neither calibration nor
objective, and $\rstar\perp\nullsp(\widetilde S)$, so the optimizer set is
$\rstar+\nullsp(\widetilde S)$. \by{$\widetilde S q=0$}
\item $\chi_B=0$ iff some calibrated $r$ lies in $\nullsp(W_B^\top)$, which
holds iff $w_s$ has a nonzero component outside $\spanop(W_B)$. \by{part (b)}
\item Let $N=P_{\spanop(W_B)^\perp}$ and $Nw_s\ne0$. Every invariant reader lies
in $\range(N)$ with $r^\top Nw_s=1$, so $1\le\norm{r}\,\norm{Nw_s}$, with
equality at $r=Nw_s/\norm{Nw_s}^2=\rinv$. \by{Cauchy-Schwarz}
\item For unit $w_s$, $\norm{Nw_s}=\sin\theta_{s,B}$, so
$\norm{\rinv}=1/\sin\theta_{s,B}$. \by{definition of the principal angle}
\end{claimproof}

The theorem separates current and intrinsic exposure: for every calibrated
deployed reader $u$,
\begin{equation}
0\le \chi_B\le \rho_{u,B},
\label{eq:chirho}
\end{equation}
the gap $\rho_{u,B}^2-\chi_B^2$ is removable readout cross-talk, and the
residual $\chi_B$ is representation-intrinsic under the declared calibration
constraint. Statistical leverage measures how uniquely a column is
represented by a span \citep{ordozgoiti2022generalized}; the duality
$\chi_B^2=1/\ell_B-1$ gives that classical quantity an operational safety
meaning.

Feasibility and stability are different questions. If endpoint reconstruction
errors have norm at most $\eta$, an exactly block-invariant reader still incurs
residual pair drift up to
\begin{equation}
2\eta\norm{\rinv}=\frac{2\eta}{\sin\theta_{s,B}},
\label{eq:noiseamplification}
\end{equation}
so a small quotient angle makes exact cancellation brittle: the invariant
reader exists algebraically but amplifies residual noise and estimation error
by $1/\sin\theta_{s,B}$. This is the third regime of the triage in
\cref{fig:taxonomy}, and it is invisible to both $\rho_{u,B}$ and $\chi_B$
alone.

One scalar writer is a modelling choice, and a contested one: gradient-based
search finds several independent refusal directions and multi-dimensional
concept cones rather than one direction, and orthogonality between them does
not imply independence under intervention \citep{wollschlager2025geometry}. The
diagnosis is not tied to the scalar case. For a safety mechanism with $k$
calibrated outputs spanning $\col(A)$, the multi-output form
\eqref{eq:multioutput} replaces $\chi_B^2$ by $\tr(H_A^{-1})-k$ and replaces
the angle condition by the rank condition
$\rank(P_{\spanop(W_B)^\perp}A)=k$: exact simultaneous invariance is feasible
exactly when the cone keeps full rank off the nuisance span. The three regimes
are then read off that rank and the norm of the optimal reader system, so a
cone-valued refusal mechanism is diagnosed by the same two questions as a
single direction.

\subsection{The three regimes, matched and measured}

\begin{figure}[t]
\centering
\begin{subfigure}[t]{0.48\linewidth}
\centering
\resizebox{\linewidth}{!}{%
\begin{tikzpicture}[font=\scriptsize, scale=0.86,
  arr/.style={-{Stealth[length=4pt]}, very thick}]
\fill[DeckGray!12] (-1.9,0) -- (1.9,0) -- (2.5,0.75) -- (-1.3,0.75) -- cycle;
\node[DeckGray] at (1.7,0.30) {$\spanop(W_B)$};
\draw[arr, DeckTeal] (0,0.30) -- (0.15,2.15) node[above]{$w_s$ (stable)};
\draw[arr, DeckOrange] (0,0.30) -- (1.75,0.95) node[above right=-3pt]{$w_s$ (ill-cond.)};
\draw[arr, DeckRed] (0,0.30) -- (-1.55,0.42) node[above left=-3pt]{$w_s$ (intrinsic)};
\draw[DeckOrange] (0.68,0.42) arc (18:36:0.72) node[midway, right=1pt]{$\theta$};
\node[align=center, text width=13em] at (0.3,-0.75)
  {$\norm{\rinv}=1/\sin\theta_{s,B}$};
\end{tikzpicture}%
}
\caption{Repairability is the angle to the nuisance span.}
\end{subfigure}\hfill
\begin{subfigure}[t]{0.48\linewidth}
\centering
\resizebox{\linewidth}{!}{%
\begin{tikzpicture}[font=\scriptsize, scale=0.86,
  arr/.style={-{Stealth[length=4pt]}, very thick}]
\draw[DeckNavy, thick] (-1.8,1.35) -- (2.0,0.45)
  node[pos=0.13, above=1pt]{$\{r:\ip{r}{w_s}=1\}$};
\foreach \s in {0.38,0.72,1.06}{\draw[DeckTeal!50, rotate=-12] (0.15,0.28) ellipse ({1.5*\s} and {0.62*\s});}
\fill[DeckTeal!90!black] (0.63,0.72) circle (1.7pt) node[right=2pt]{$\rstar$ (level $\chi_B$)};
\fill[DeckBlue] (-1.35,1.24) circle (1.7pt) node[above=1pt]{$u$};
\draw[DeckPurple, very thick, dotted] (-1.35,1.24) .. controls (-0.4,1.02) and (0.1,0.88) .. (0.63,0.72);
\node[DeckPurple, anchor=north, align=center] at (-0.75,-0.05) {ridge path $\rlam$};
\end{tikzpicture}%
}
\caption{Calibrated readers: level sets of $\norm{W_B^\top r}$ meet the
constraint at $\rstar$.}
\end{subfigure}
\caption{The framework in linear algebra. (a) The three regimes are angles of
$w_s$ to the nuisance span. (b) Calibrated readers: the intrinsic exposure
$\chi_B$ is the smallest level set of $\norm{W_B^\top r}$ touching the
calibration constraint, and the regularized reader $\rlam$ of \cref{thm:ridge}
traces the path from the deployed head $u$ to $\rstar$.}
\label{fig:geometry}
\end{figure}
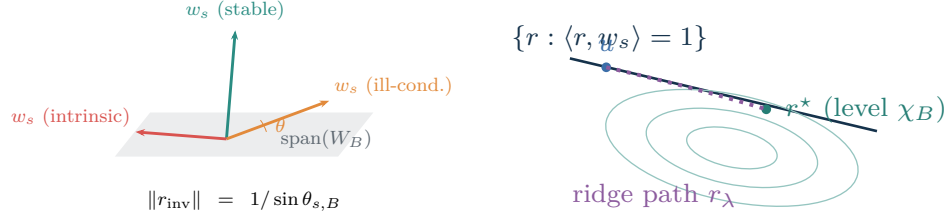

\Cref{fig:geometry} draws the complete diagnosis. The angle between $w_s$ and
$\spanop(W_B)$ determines everything: a right angle is the stable removable
regime, a small positive angle is removable but ill-conditioned, and an
in-span safety writer is the intrinsic collision, where
\cref{thm:leverage}(b) denies the existence of any invariant calibrated
reader.

The next construction shows the diagnosis is invisible to current
exposure alone.

\begin{proposition}[Same current exposure, opposite repairability]
\label{prop:matched}
Let $L$ be even, $d\ge L+1$, $w_s=e_0$, $0<\beta<1$, and
$\gamma=\sqrt{1-\beta^2}$. The removable block
$b_j=\beta e_0+\gamma e_j$, $j=1,\dots,L$, and the intrinsic paired block
\begin{equation*}
b_{2k-1}=\beta e_0+\gamma e_k,\qquad
b_{2k}=\beta e_0-\gamma e_k,\qquad k=1,\dots,L/2,
\end{equation*}
have the same tied exposure $\rho=\beta\sqrt L$. For the first, $\chi_B=0$ and
$\norm{\rinv}^2=1+L\beta^2/(1-\beta^2)$. For the second, $\chi_B=\rho$ and no
exactly invariant calibrated reader exists.
\end{proposition}

\begin{claimproof}
\item Both blocks give $W_B^\top e_0=\beta\mathbf 1_L$, so
$\rho=\norm{W_B^\top e_0}=\beta\sqrt L$ in both cases. \by{unit columns}
\item For the removable block, $r=e_0-(\beta/\gamma)\sum_{j=1}^Le_j$ satisfies
$r^\top e_0=1$ and $r^\top b_j=\beta-(\beta/\gamma)\gamma=0$, so $\chi_B=0$.
\by{explicit annihilating reader}
\item Its squared norm is $1+L\beta^2/\gamma^2$, and it equals $\rinv$ because
it is the calibrated element of $\spanop(W_B)^\perp$ of minimum norm.
\by{\cref{thm:leverage}(c)}
\item For the paired block, any calibrated $r$ has $r^\top e_0=1$, so writing
$r_k$ for its $e_k$ coordinates,
$\norm{W_B^\top r}^2=\sum_{k=1}^{L/2}[(\beta+\gamma r_k)^2+(\beta-\gamma r_k)^2]
=L\beta^2+2\gamma^2\sum_kr_k^2\ge L\beta^2$. \by{cross terms cancel}
\item The tied reader $e_0$ attains the bound, so $\chi_B=\beta\sqrt L=\rho$,
and $\chi_B>0$ denies exact invariance. \by{\cref{thm:leverage}(b)}
\end{claimproof}

\begin{figure}[tbp]
\centering
\includegraphics[width=0.98\linewidth]{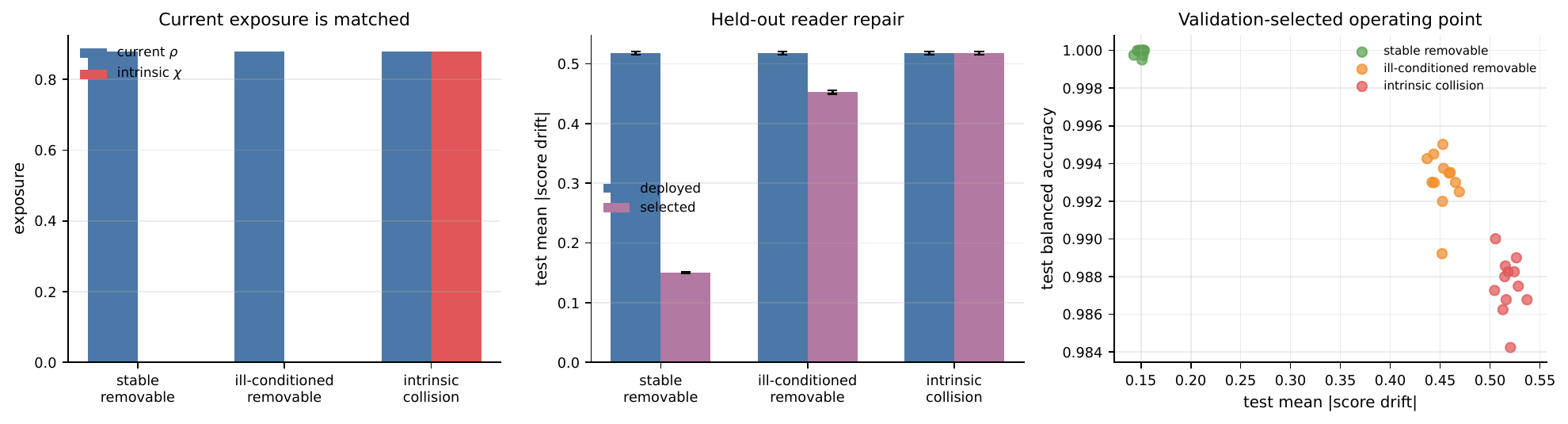}
\caption{Matched-current-exposure benchmark. Left: the three families share
$\rho=\MatchedRho$ but have intrinsic exposure
$\chi=\StableChi/\IllChi/\IntrinsicChi$. Middle: held-out drift before and
after validation-selected reader repair. Right: the selected operating points, where the two removable families trade
drift for accuracy and the intrinsic family holds its drift.}
\label{fig:matched}
\end{figure}

\begin{figure}[tbp]
\centering
\includegraphics[width=0.98\linewidth]{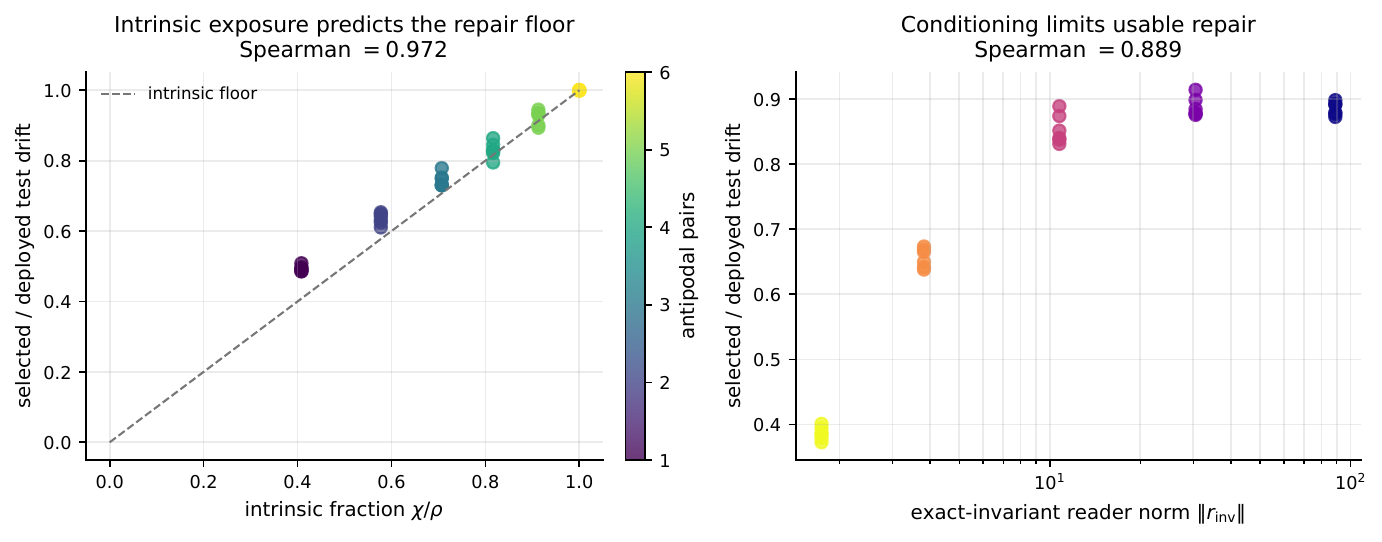}
\caption{Out-of-sample repairability across $\SweepGeometries$ independently
generated geometries. Left: the intrinsic fraction $\chi/\rho$ predicts the
held-out residual drift ratio (Spearman $\IntrinsicSweepSpearman$); the dashed
diagonal is the intrinsic floor. Right: on the removable subset, the
exact-invariant reader norm predicts how much repair survives the utility
constraint (Spearman $\ConditioningSweepSpearman$).}
\label{fig:sweep}
\end{figure}

Equal observed tied-head cross-talk can therefore represent a completely
removable problem or a completely intrinsic one; the numerical audit adds a
third, ill-conditioned removable family whose $\chi_B$ is zero but whose
$\norm{\rinv}$ is large, and confirms the full three-way diagnosis on held-out
fibers (\cref{tab:matched}, \cref{fig:matched}) and across independently
generated geometries (\cref{fig:sweep}).

The matched benchmark (\cref{tab:matched}, \cref{fig:matched}) holds current exposure fixed at
$\rho=\MatchedRho$ with sample-level deployed scores matched across families to
$\DeployedScoreMismatch$, so held-out differences are attributable to geometry
alone. A behavioral disagreement rate is the primary observable of black-box
multilingual red-teaming on deployed models
\citep{yong2024lowresource,deng2024multilingual}; the three families are
constructed to be indistinguishable to that observable, and the held-out
columns of \cref{tab:matched} show what it leaves undetermined.
\citet{arditi2024refusal} identify a refusal direction that is causally
necessary and sufficient in English prompts, which establishes that a linear
read of safety exists; whether a calibrated re-read can escape the language
block is exactly the quantity $\chi$ measures. Validation-selected readers reduce held-out drift by
$\StableDriftReduction\%$ in the stable removable family but only
$\IllDriftReduction\%$ in the ill-conditioned removable family, and
$\IntrinsicDriftReduction\%$ in the intrinsic collision, matching
$\chi=\StableChi/\IllChi/\IntrinsicChi$ and
$\norm{\rinv}=\StableInvariantNorm/\IllInvariantNorm$. The exact invariant
reader illustrates the conditioning regime directly: it reaches
$\StableExactBacc\%$ balanced accuracy in the stable family but collapses to
$\IllExactBacc\%$ in the ill-conditioned one, because its norm
$\IllInvariantNorm$ amplifies endpoint noise
(\cref{eq:noiseamplification}). The collapse is quantitative as well as
directional: with endpoint residual scale $0.10$, the amplification bound
evaluates to $2\times 0.10\times\IllInvariantNorm\approx 18$ for the
ill-conditioned family, more than twenty times the generator's signal scale
$0.78$, while the stable family's $2\times 0.10\times\StableInvariantNorm
\approx 0.27$ stays comfortably below it; the observed accuracies are what
this arithmetic requires. The conditioning numbers also close a loop with the
budget comparison of \cref{sec:measure-certify}: for the unit tied writer,
$\kappa=\cos\theta_{s,B}$ while $\norm{\rinv}=1/\sin\theta_{s,B}$, so
$\norm{\rinv}=1/\sqrt{1-\kappa^2}$, and the two stages, computing these
quantities by different formulas on different code paths, agree to four
significant figures in both removable families ($\BudgetStableKappa$ gives
$1.35$ against $\StableInvariantNorm$; the ill-conditioned $\kappa$, within
one part in ten thousand of the ceiling, gives $91.6$ against
$\IllInvariantNorm$). The intrinsic family closes the pattern: $\kappa$ at
the ceiling means $\sin\theta_{s,B}=0$ and no invariant reader at any norm.
The regularization frontiers behind this
table are in \cref{fig:frontiers} (appendix).

\begin{table}[t]
\centering
\small
\caption{Matched-current-exposure benchmark.  All three families have identical
sample-level deployed scores.  Readers are estimated from training contrasts,
selected on validation fibers under a one-percentage-point utility constraint,
and evaluated once on test fibers.  BAcc is balanced accuracy; ``exact BAcc''
uses the minimum-norm exactly invariant reader when it exists.}
\label{tab:matched}
\resizebox{\linewidth}{!}{%
\begin{tabular}{lrrrrrrr}
\toprule
Geometry & $\rho$ & $\chi$ & $\|r_{\mathrm{inv}}\|$ & deployed drift & selected drift & selected BAcc & exact BAcc \\
\midrule
Stable removable & \MatchedRho & \StableChi & \StableInvariantNorm & \StableDeployedDrift & \StableSelectedDrift & \StableSelectedBacc & \StableExactBacc \\
Ill-conditioned removable & \MatchedRho & \IllChi & \IllInvariantNorm & \IllDeployedDrift & \IllSelectedDrift & \IllSelectedBacc & \IllExactBacc \\
Intrinsic collision & \MatchedRho & \IntrinsicChi & n/a & \IntrinsicDeployedDrift & \IntrinsicSelectedDrift & \IntrinsicSelectedBacc & n/a \\
\bottomrule
\end{tabular}%
}
\end{table}

\Cref{fig:matched} shows the same separation graphically: the exposure
bars are indistinguishable across families, the held-out drift bars are
not, and the operating-point panel isolates the diagnostic signature of
each regime; only the intrinsic family sits pinned at its deployed drift
with nothing to trade.

Across $\SweepGeometries$ independently rotated geometries
(\cref{fig:sweep}), the intrinsic fraction $\chi/\rho$ predicts the residual
drift floor with Spearman correlation $\IntrinsicSweepSpearman$ (ratios rising
from $\IntrinsicLowResidualRatio$ at one antipodal pair to
$\IntrinsicHighResidualRatio$ at a fully paired block), and on the removable
subset $\log\norm{\rinv}$ predicts usable repair with Spearman
$\ConditioningSweepSpearman$ (ratios from $\StableSweepResidualRatio$ to
$\IllSweepResidualRatio$). The two panels of \cref{fig:sweep} are one
decomposition read twice: by \cref{eq:chirho} the diagonal $\chi/\rho$ is the
non-negotiable floor, so a geometry's height above the diagonal is drift that
is removable in principle but unrecovered under the utility constraint; on
the removable subset, where the floor is identically zero, that height is the
entire residual and climbs monotonically as the invariant-reader norm climbs
by an order of magnitude. The floor explains position along the diagonal, the
conditioning explains height above it, and nothing else is left to explain. Finite-sample learning curves
(\cref{fig:learning}, appendix) sweep the audit sample from
$\LearningMinTrain$ to $\LearningMaxTrain$ fibers: the stable ratio falls from
$\StableRatioSixteen$ to $\StableRatioMax$, the ill-conditioned ratio from
$\IllRatioSixteen$ to $\IllRatioMax$, and the intrinsic ratio stays at
$\IntrinsicRatioMax$; more data sharpens the estimate of the quotient geometry the representation
already carries.
The plateaus land where the matched benchmark says they must: at
$\LearningMaxTrain$ fibers the residual ratios $\StableRatioMax$,
$\IllRatioMax$, and $\IntrinsicRatioMax$ sit near one minus the matched
drift reductions for the same three families, from independent seeds, sample
sizes, and selection procedures. Estimation noise is what vanishes with data;
the geometric remainder is what two independent experiments jointly locate.

\section{The causal interface}
\label{sec:test}

Define the calibrated feature setter
\begin{equation}
T_i^a(h)=h+\bigl(a-r_i^\top h\bigr)w_i,
\label{eq:setter}
\end{equation}
which writes along $w_i$ until the reading along $r_i$ equals $a$; calibration
$r_i^\top w_i=1$ guarantees $r_i^\top T_i^a(h)=a$. This is an
activation-patching intervention with the read and write directions allowed to
differ, and separate readers and writers make the interaction directional.

\begin{theorem}[Untied order-swap identity]
\label{thm:untied}
Let $C=R^\top W$, let $i\ne j$, and let $w_i,w_j\ne0$. Then
\begin{equation}
T_i^aT_j^b(h)-T_j^bT_i^a(h)
=\bigl(a-r_i^\top h\bigr)C_{ji}w_j-\bigl(b-r_j^\top h\bigr)C_{ij}w_i.
\label{eq:untiedcommutator}
\end{equation}
The setters commute for all $h,a,b$ if and only if $C_{ij}=C_{ji}=0$. If the
target increments are independent, centered, and have variance $\sigma^2$, and
the writers are unit norm, then
$\E\norm{T_i^aT_j^b(h)-T_j^bT_i^a(h)}^2=\sigma^2(C_{ij}^2+C_{ji}^2)$. A
one-sided target change identifies either directional entry.
\end{theorem}

\begin{claimproof}
\item Write $\delta_i=a-r_i^\top h$ and $\delta_j=b-r_j^\top h$. \by{notation}
\item $T_j^b(h)=h+\delta_jw_j$ has $i$-reading $r_i^\top h+\delta_jC_{ij}$.
\by{$r_i^\top w_j=C_{ij}$}
\item $T_i^aT_j^b(h)=h+\delta_jw_j+(\delta_i-\delta_jC_{ij})w_i$.
\by{apply \eqref{eq:setter}}
\item Symmetrically, $T_j^bT_i^a(h)=h+\delta_iw_i+(\delta_j-\delta_iC_{ji})w_j$;
subtracting gives \eqref{eq:untiedcommutator}. \by{cancel $h$}
\item If $C_{ij}=C_{ji}=0$ the defect vanishes identically; conversely,
choosing $\delta_j=0$, $\delta_i=1$ makes the defect $C_{ji}w_j$, and
$\delta_i=0$, $\delta_j=1$ makes it $-C_{ij}w_i$, so commutation for all
targets forces $C_{ji}=C_{ij}=0$ since $w_i,w_j\ne0$. \by{one-sided choices}
\item With independent centered increments of variance $\sigma^2$ and unit
writers, the cross term has zero mean and
$\E\norm{\text{defect}}^2=\sigma^2(C_{ji}^2\norm{w_j}^2+C_{ij}^2\norm{w_i}^2)
=\sigma^2(C_{ij}^2+C_{ji}^2)$. \by{independence}
\item The same one-sided choices as step (5) leave a scalar multiple of one
known writer, identifying the corresponding directional entry.
\by{read off \eqref{eq:untiedcommutator}}
\end{claimproof}

\begin{figure}[tbp]
\centering
\includegraphics[width=0.98\linewidth]{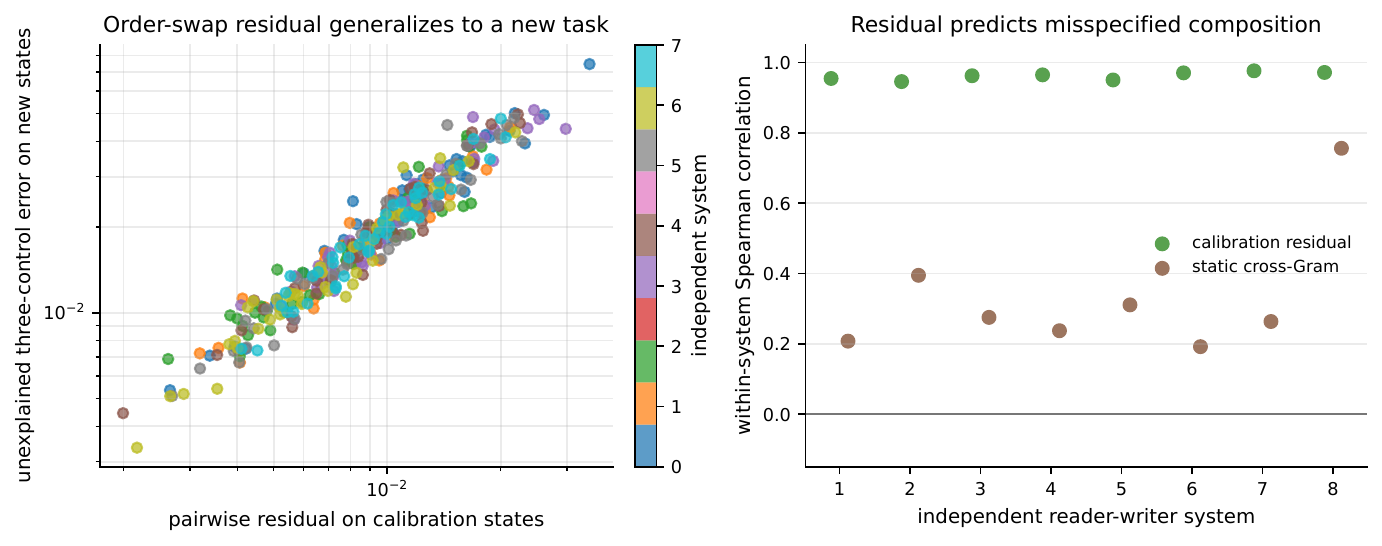}
\caption{Untied diagnostic under nonlinear, state-dependent setters. Left: the
pairwise order-swap residual on calibration states predicts the unexplained
three-control composition error on new states. Right: within-system Spearman
correlations for the residual against the static cross-Gram baseline.}
\label{fig:untied}
\end{figure}

The value of \cref{thm:untied} is operational. When $R$ and $W$ are available,
$C$ is directly computable, and explicit interventions test whether the learned
controls behave like the proposed linear reader/writer interface. For an
observed order difference $\Delta^{\mathrm{obs}}_{ij}$, define the residual
\begin{equation}
\calR_{ij}=\Delta^{\mathrm{obs}}_{ij}
-\bigl[(a-r_i^\top h)C_{ji}w_j-(b-r_j^\top h)C_{ij}w_i\bigr].
\label{eq:commresidual}
\end{equation}
A large residual localizes nonlinear leakage, state dependence, off-dictionary
components, or intervention implementation error; the diagnostic content lives
in the residual, not in the predicted linear term. This is the
consistency-test logic of causal abstraction \citep{geiger2025causal},
instantiated for a linear read/write interface. The empirical literature on
steering supplies the motive: steering vectors are unreliable in and out of
distribution, with variance across inputs traced to the geometry of activation
differences and to whether one direction is coherent at all
\citep{tan2024steering,braun2025unreliability}, and reliability degrades
further under multi-attribute control, which has prompted replacing a single
static vector by a state-dependent field \citep{li2026steeringfields}. Those
studies measure that composition fails; \cref{eq:commresidual} is a quantity
computed before the composition is attempted, and the audit below asks it to
predict the failure out of sample.
The closing audit of this
section estimates the residual on calibration states and asks it to predict a
distinct three-control composition error on new states and targets
(\cref{fig:untied}).

\begin{corollary}[Tied specialization and signed one-sided defect]
\label{cor:tied}
For the tied system $R=W$ with unit columns, \eqref{eq:untiedcommutator}
becomes $T_i^aT_j^b(h)-T_j^bT_i^a(h)=g_{ij}[(a-r_i)w_j-(b-r_j)w_i]$ with
$g_{ij}=\ip{w_i}{w_j}$ and readings $r_i=\ip{w_i}{h}$, so the maps commute for
all $a,b,h$ iff $g_{ij}=0$, and independent centered targets of variance
$\sigma^2$ give expected squared defect $2\sigma^2g_{ij}^2$. With $b=r_j$ and
$a=r_i+1$ the defect is $g_{ij}w_j$, so
$g_{ij}=\ip{T_i^aT_j^b(h)-T_j^bT_i^a(h)}{w_j}$, recovering the signed entry.
\end{corollary}

\begin{claimproof}
\item Set $R=W$ in \cref{thm:untied}: $C_{ij}=C_{ji}=g_{ij}$ and
$\delta_i=a-r_i$, $\delta_j=b-r_j$. \by{tied case}
\item The defect becomes $g_{ij}[(a-r_i)w_j-(b-r_j)w_i]$ and the variance
formula gives $\sigma^2\cdot 2g_{ij}^2$. \by{\cref{thm:untied}}
\item $b=r_j$, $a=r_i+1$ leaves $g_{ij}w_j$; its inner product with the unit
vector $w_j$ is $g_{ij}$. \by{one-sided choice}
\end{claimproof}

Because $g_{ij}=w_i^\top w_j$ is directly computable when $W$ is known, the
tied identity's operational content is a consistency check rather than a method
of Gram estimation: the residual \eqref{eq:commresidual} is zero under the
ideal tied read/write map and nonzero under untied read/write, state
dependence, nonlinearity, off-dictionary components, or implementation error.
When the data are generated by that identity, the squared-defect average is a
Monte Carlo check of it rather than independent causal evidence.

The stage audit instantiates both regimes. For the tied system, across $\CommDistinctPairs$ distinct feature pairs with
$\CommInterventions$ interventions each, the explicit two-order composition
matches the closed form to $\CommIdentityError$, two units in the last place
of double precision, so the check saturates the arithmetic itself; the
squared-coherence estimate
has correlation $\CommCorr$ and RMSE $\CommRmse$, aggregation gives a tied-head
block norm of $\BlockRhoEstimate$ against the true $\BlockRhoTrue$, and the
signed one-sided recovery of \cref{cor:tied} has correlation $\SignedCommCorr$
and RMSE $\SignedCommRmse$ (\cref{fig:tiedcomm}, appendix). These
Monte Carlo recoveries confirm internal consistency of the tied-map identity
rather than supplying independent evidence for it. They do close a loop
between statics and dynamics: the number the measuring audit reads off $W$ in
one matrix multiplication, the block norm $\BlockRhoTrue$, is reconstructed
purely from the order dependence of interventions as $\BlockRhoEstimate$, and
the signed protocol recovers individual Gram entries at correlation
$\SignedCommCorr$. On synthetic data this is consistency; on a real stack it
is the template for estimating $\rho$ where $W$ is not given.

The untied diagnostic is the out-of-sample headline
(\cref{fig:untied}). Across $\UntiedSystems$ independent reader-writer systems, the regime of
learned encoder/decoder pairs whose read and write directions differ by
construction \citep{cunningham2023sae,bricken2023monosemanticity}, with
actual setters carrying nonlinear, state-dependent leakage, the
pairwise order-swap residual \eqref{eq:commresidual}, estimated on calibration
states, predicts the unexplained part of a three-control composition error on
new states and targets with median within-system Spearman correlation
$\UntiedMedianResidualSpearman$ (pooled $\UntiedPooledResidualSpearman$ over
$\UntiedTriples$ held-out triples), against
$\UntiedMedianCrossGramSpearman$ (pooled $\UntiedPooledCrossGramSpearman$) for
the static cross-Gram magnitude alone. The simulated residual, not the static
linear score, carries the transferable information about how controls
miscompose.

\section{Repair and limits}

Two repairs answer two situations: designing a new calibrated reader when the
contrast geometry (or its covariance) is available, and editing the head you
have when only audited pairs are. Both act on the reader side of a fixed
representation, which places them against two established families:
domain-adversarial training and invariant risk minimization make a nuisance
attribute non-discriminable, or make the optimal readout invariant to it, by
training against a learned classifier \citep{ganin2016dann,arjovsky2019irm},
and closed-form linear concept erasure removes a labeled attribute at the
data level against every classifier and norm \citep{belrose2023leace}, a line
that began with adversarial and kernelized erasure of a concept subspace
\citep{ravfogel2022linear,ravfogel2022kernelized}. Erasure asks that no linear
predictor recover the attribute; \cref{thm:drift}(b) is the fiber-conditional
form of that condition, imposed on one declared score rather than on every
predictor. The decomposition here is calibrated and block-specific and emits a
verdict no erasure method computes: \cref{thm:projection} edits the deployed
head with a held-out transfer bound, \cref{thm:ridge} designs a new reader
along an explicit exposure-conditioning frontier, and $\chi_B$
(\cref{def:chi}) states when no head-side object of either kind can succeed. The section closes with
what no reader optimization can achieve.

\subsection{Two closed-form repairs}

\begin{proposition}[Regularized minimum-variance reader]
\label{thm:ridge}
For $\lambda>0$, the problem
$\min_{r^\top w_s=1}\{\norm{W_B^\top r}^2+\lambda\norm{r}^2\}$
has the unique solution
\begin{equation}
\rlam=\frac{(W_BW_B^\top+\lambda\Id)^{-1}w_s}
{w_s^\top(W_BW_B^\top+\lambda\Id)^{-1}w_s},
\label{eq:rlambda}
\end{equation}
with minimum value $[w_s^\top(W_BW_B^\top+\lambda\Id)^{-1}w_s]^{-1}$. If
$c\sim(0,\Id)$ and independent residual noise has covariance $\lambda\Id$, the
objective is exactly the expected squared score drift.
\end{proposition}

\begin{claimproof}
\item $Q_\lambda=W_BW_B^\top+\lambda\Id\succ0$, and for calibrated $r$,
$r^\top Q_\lambda r=\norm{W_B^\top r}^2+\lambda\norm{r}^2$. \by{expand}
\item $1=(r^\top w_s)^2\le(r^\top Q_\lambda r)(w_s^\top Q_\lambda^{-1}w_s)$.
\by{Cauchy-Schwarz in the $Q_\lambda$ metric}
\item Equality holds iff $Q_\lambda r\propto w_s$, which after calibration is
exactly \eqref{eq:rlambda}; strict convexity gives uniqueness.
\by{equality case}
\item $\E(r^\top(W_Bc+e))^2=\norm{W_B^\top r}^2+\lambda\norm{r}^2$ when
$\Cov(c)=\Id$, $\Cov(e)=\lambda\Id$, independent. \by{expand the square}
\end{claimproof}

As $\lambda\to\infty$, $\rlam$ approaches the normalized tied reader
$w_s/\norm{w_s}^2$; as $\lambda\downarrow0$ it approaches the minimum-norm
minimum-exposure reader, which in the exactly removable case is $\rinv$. Hence
$\lambda$ traces a continuous frontier between current readout behavior and
exact nuisance cancellation, the dotted path of \cref{fig:geometry}(b). How a
reader selected on audited pairs then behaves on unseen pairs is a separate,
statistical question, treated in a companion manuscript
\citep{ahnouch2026interpolation}; here $\lambda$ is a design knob on the
frontier and the audits below report one frozen test evaluation of it. A
safety mechanism with several calibrated outputs $A\in\R^{d\times k}$ (full
column rank) admits the same construction: with
$S_A=AA^\top+W_BW_B^\top$ and $H_A=A^\top S_A^\dagger A$,
\begin{equation}
\min_{R_A^\top A=\Id_k}\norm{W_B^\top R_A}_F^2=\tr(H_A^{-1})-k,\qquad
R_A^\star=S_A^\dagger AH_A^{-1},
\label{eq:multioutput}
\end{equation}
and exact simultaneous invariance is feasible precisely when
$\rank(P_{\spanop(W_B)^\perp}A)=k$; the claimproof is in
the proof below. This covers multi-category safety heads without
forcing every policy dimension into one scalar direction.

\begin{claimproof}
\item For $R^\top A=\Id_k$,
$\tr(R^\top S_AR)=\tr(R^\top AA^\top R)+\norm{W_B^\top R}_F^2
=k+\norm{W_B^\top R}_F^2$. \by{expand $S_A$}
\item Minimizing $\tr(R^\top S_AR)$ under the matrix constraint has
least-energy solution $R^\star=S_A^\dagger A(A^\top S_A^\dagger A)^{-1}
=S_A^\dagger AH_A^{-1}$. \by{generalized least squares}
\item $R^\star$ is feasible: $R^{\star\top}A=H_A^{-1}A^\top S_A^\dagger A
=\Id_k$, using $\range(A)\subseteq\range(S_A)$. \by{substitute}
\item Substitution gives the minimum $\tr(H_A^{-1})$, hence
\eqref{eq:multioutput}. \by{steps (1)-(3)}
\item Exact simultaneous invariance solves
$R^\top P_{\spanop(W_B)^\perp}A=\Id_k$, which is solvable iff the projected
anchor matrix has full column rank $k$. \by{surjectivity}
\end{claimproof}

For repair from audited pairs alone, let $D$ be audited pairs with hidden
contrast span $\calL_D=\spanop\{h(x)-h(x'):(x,x')\in D\}$.

\begin{proposition}[Least-change projection and its held-out bound]
\label{thm:projection}
$u_D=(\Id-P_{\calL_D})u$ is the unique minimizer of $\norm{v-u}$ over
$\calL_D^\perp$, and $u_D^\top(h(x)-h(x'))=0$ for all $(x,x')\in D$. For any
held-out hidden contrast $d$,
$\;|u_D^\top d|\le\norm{u_D}\,\norm{(\Id-P_{\calL_D})d}$.
\end{proposition}

\begin{claimproof}
\item Orthogonal split $u=P_{\calL_D}u+(\Id-P_{\calL_D})u$ gives
$\norm{u-v}^2=\norm{P_{\calL_D}u}^2+\norm{(\Id-P_{\calL_D})u-v}^2$ for
$v\in\calL_D^\perp$, minimized at $v=u_D$. \by{Pythagoras}
\item $u_D\in\calL_D^\perp$ annihilates every training contrast. \by{definition}
\item $u_D^\top d=u_D^\top(\Id-P_{\calL_D})d$, so
$|u_D^\top d|\le\norm{u_D}\,\norm{(\Id-P_{\calL_D})d}$. \by{Cauchy-Schwarz}
\end{claimproof}

Generalization thus depends on how well a new contrast lies in the training span;
an empirical basis $\widehat\calL_{D,0.995}$ that retains a fixed energy fraction
is a denoised truncation of $\calL_D$, not the exact span. In practice one also
solves the penalized problem
$\min_{u,b}\sum_k(u^\top h_k+b-y_k)^2
+\lambda\sum_{D}(u^\top(h(x)-h(x')))^2+\gamma\norm{u}^2$, with closed-form
solution $\widehat\beta=(X^\top X+\lambda D^\top D+\gamma R)^\dagger X^\top y$
($R=\diag(\Id_d,0)$). The scale-dependent $\lambda$ must be selected
on validation data under a prespecified utility constraint, with the head, bias,
threshold, and $\lambda$ frozen before the single test evaluation.

\begin{figure}[tbp]
\centering
\begin{minipage}{0.48\linewidth}\centering
\includegraphics[width=\linewidth]{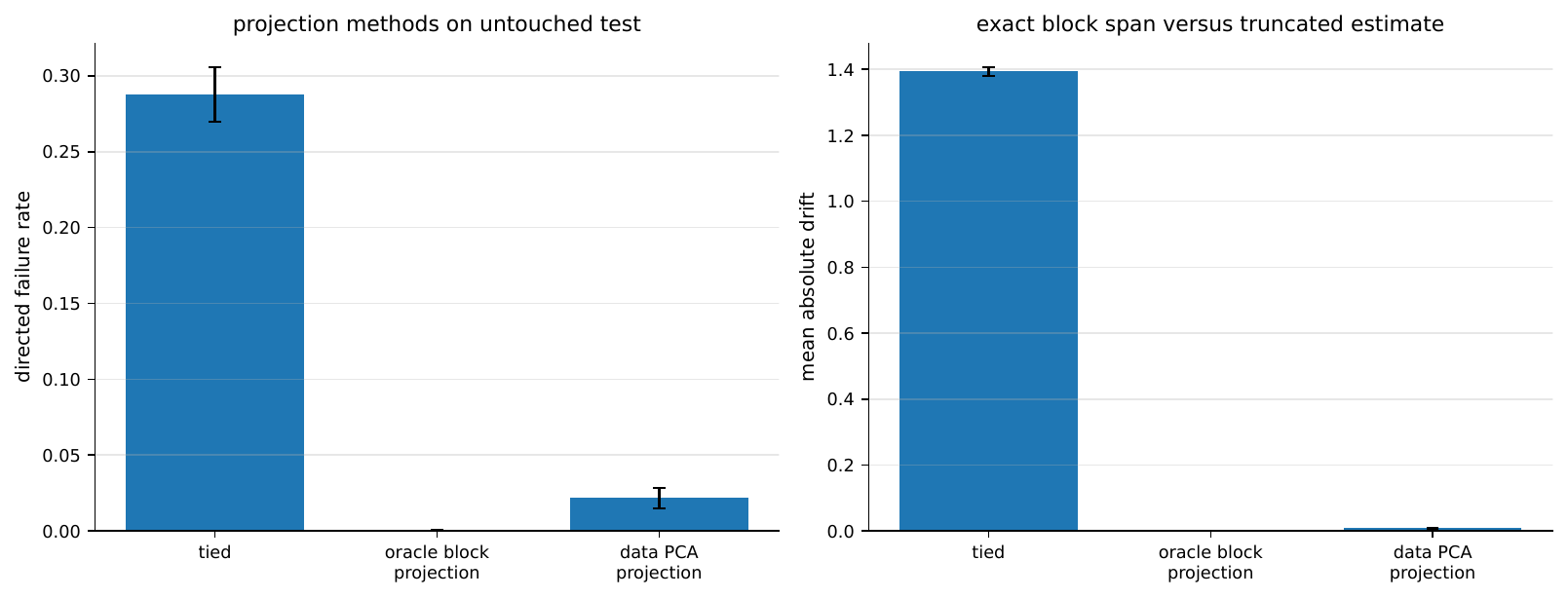}\end{minipage}\hfill
\begin{minipage}{0.48\linewidth}\centering
\includegraphics[width=\linewidth]{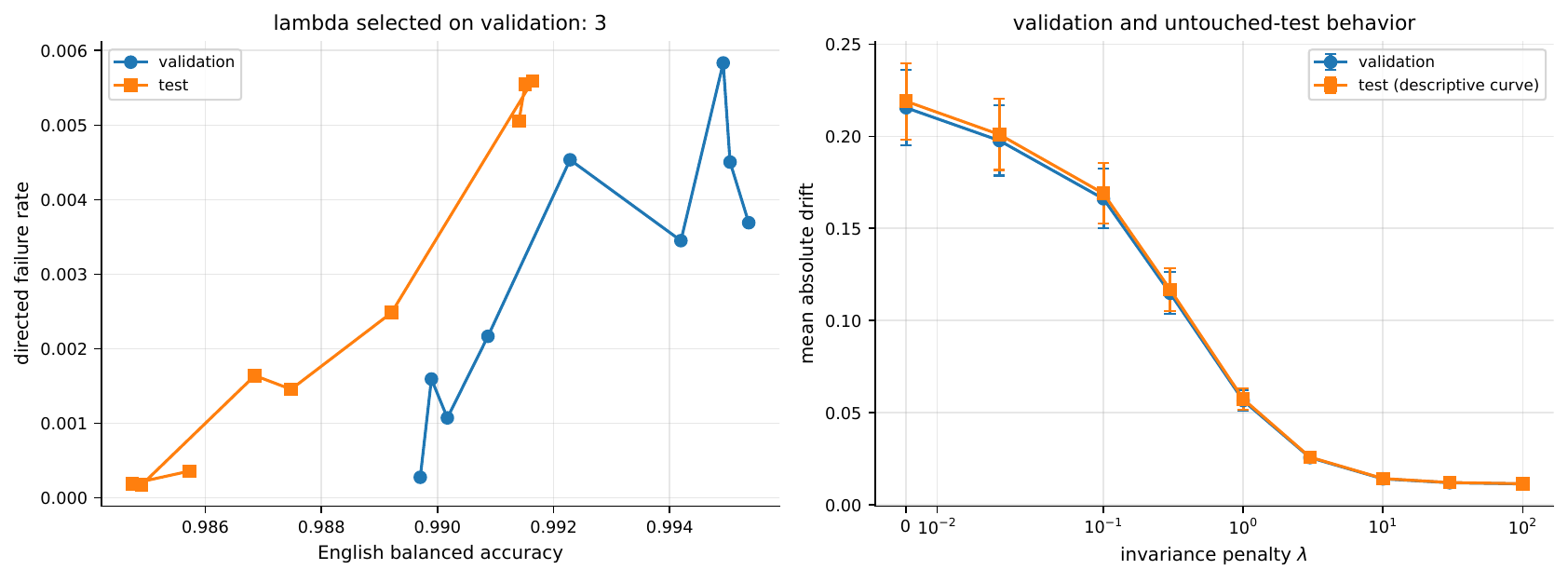}\end{minipage}
\caption{Left: exact oracle block-span projection removes almost all held-out
drift, while the data-driven truncated-PCA projection is a denoised approximation.
Right: the invariant-ridge frontier on validation and untouched test; $\lambda$ is
selected on validation.}
\label{fig:repair}
\end{figure}

\label{sec:whichrepair}
The two constructions solve different problems and require different inputs.
The regularized reader $\rlam$ needs the block geometry $W_B$ or, in its
dictionary-free form, the empirical contrast covariance
$\widehat\Sigma_\Delta$; it designs the best calibrated reader outright and
inherits the diagnosis of \cref{thm:leverage}: its frontier is steep exactly
when the geometry is well conditioned. The projection $u_D$ needs only audited
pairs, changes the deployed head as little as possible in Euclidean norm, and
carries the held-out transfer bound of \cref{thm:projection}. When
$\chi_B>0$, neither construction removes the block response, which is the
point of the diagnosis: repair budgets should go to the representation, not
the head. Both repairs are evaluated on untouched test fibers in
\cref{fig:repair}.

\Cref{fig:repair} shows the head-repair results. On the untouched test split,
empirical projection lowers the tied head's mean absolute drift from
$\TiedTestDrift$ to $\OracleProjTestDrift$ (oracle block span) and
$\TruncProjTestDrift$ (data PCA truncation, rank $\TruncBasisRank$), and the
directed unsafe rate from $\TiedTestFailurePct\%$ to
$\OracleProjTestFailurePct\%$ and $\TruncProjTestFailurePct\%$ respectively; at
the fixed threshold $\tau=0.65$ the tied rate is $\TiedFixedTauFailurePct\%$
against $\OracleFixedTauFailurePct\%$ and $\TruncFixedTauFailurePct\%$. The
fixed and recalibrated columns decompose the gain: recalibration alone moves
the tied head by about four points, and the projection removes essentially
everything that remains because it removes the drift itself, three orders of
magnitude in mean absolute drift; the truncated projection's small residual
under recalibration is an operating-point choice trading margin for accuracy,
not surviving cross-talk, since its fixed-threshold rate sits at
$\TruncFixedTauFailurePct\%$. The
oracle projection removes $\OracleRemovedHeadEnergy\%$ of the head energy
(retained head norm $\OracleProjHeadNorm$), so a small Euclidean edit can be a
large functional one. For the invariant ridge, validation selects
$\lambda=\RepairSelectedLambda$; the frozen test directed rate falls from
$\RepairBaseTestFailurePct\%$ to $\RepairSelectedTestFailurePct\%$ (Wilson
interval $[\RepairSelectedTestWilsonLow,\RepairSelectedTestWilsonHigh]\%$) and
mean absolute drift from $\RepairBaseTestDrift$ to $\RepairSelectedTestDrift$.
Because the baseline is already small, the absolute reduction is the
informative headline number, and $\lambda$ remains a validation choice rather
than a test-set optimum.

\subsection{What aggregates fix, and what they leave free}

Two global statements frame what an audit of this kind concludes. The level
leaves the structure free, and reader optimization meets a floor set by rank. Both
refine classical objects: global coherence floors on maximum cross-correlation
are fundamental \citep{welch1974lower}, and the aggregate interaction of two
frames is the subject of cross-frame potentials \citep{aceska2022crossframe};
the theorem measures what remains of both after every reader is optimized
against a fixed writer system.

\begin{theorem}[Global floors, reader-optimized capacity, and non-identification]
\label{thm:global}
Let $W\in\R^{d\times n}$ have unit columns, $r=\rank(W)$, $G=W^\top W$,
$S=WW^\top$, and leverage scores $\ell_i=w_i^\top S^\dagger w_i$.
\begin{enumerate}[label=\textup{(\alph*)},leftmargin=*,itemsep=1pt,topsep=2pt]
\item $\lambda_{\max}(G)\ge n/r$ and
$\tfrac1n\sum_{i}\sum_{j\ne i}G_{ij}^2\ge n/r-1$; the $n/d$ bounds follow from
$r\le d$.
\item Over all self-calibrated reader systems,
\begin{equation}
\min_{\diag(R^\top W)=\mathbf 1}\norm{\off(R^\top W)}_F^2
=\sum_{i=1}^n\Bigl(\frac{1}{\ell_i}-1\Bigr)\ \ge\ \frac{n(n-r)}{r},
\label{eq:globalfloor}
\end{equation}
with minimum-norm optimal readers $r_i^\star=S^\dagger w_i/\ell_i$, and
equality in the rank-only bound iff $\ell_1=\dots=\ell_n=r/n$.
\item Fix a safety index $\isafe$ and equal-cardinality sets
$B,C\not\ni\isafe$, and let $\Pi$ be a column permutation fixing $\isafe$ and
mapping $B$ onto $C$. Then $W'=W\Pi$ has $G'=\Pi^\top G\Pi$ with the same
spectrum, rank, trace, and frame potential as $G$, yet
$\rho_{\isafe,B}(W')^2=\sum_{k\in C}G_{\isafe k}^2$.
\end{enumerate}
\end{theorem}

\begin{claimproof}
\item The nonzero eigenvalues of $G$ sum to $n$ over at most $r$ of them, so
$\lambda_{\max}(G)\ge n/r$. \by{averaging}
\item $\norm{G}_F^2\ge(\tr G)^2/r=n^2/r$ and
$\norm{G}_F^2=n+\sum_{i\ne j}G_{ij}^2$. \by{rank-$r$ Frobenius bound}
\item For each calibrated reader,
$r_i^\top Sr_i=\sum_j(r_i^\top w_j)^2=1+\sum_{j\ne i}(r_i^\top w_j)^2$.
\by{expand $S$}
\item Each row minimizes independently, and \cref{thm:leverage}(a) applied to
$w_i$ against the full dictionary gives row minimum $1/\ell_i-1$ at
$r_i^\star=S^\dagger w_i/\ell_i$. \by{rows decouple}
\item $\sum_i\ell_i=\tr(W^\top S^\dagger W)=\tr(S^\dagger S)=r$, so
Cauchy-Schwarz gives $\sum_i1/\ell_i\ge n^2/r$, with equality iff all $\ell_i$
are equal to $r/n$. \by{Cauchy-Schwarz}
\item $G'=\Pi^\top G\Pi$ is an orthogonal similarity, preserving spectrum,
rank, trace, and $\norm{G}_F$. \by{permutation similarity}
\item $\Pi$ fixes $\isafe$ and sends $B$ to $C$, so
$\rho_{\isafe,B}(W')^2=\sum_{k\in C}G_{\isafe k}^2$.
\by{$G'_{\isafe j}=G_{\isafe\pi(j)}$}
\end{claimproof}

Part (c) makes non-identification exact: equal overcompleteness and equal
global coherence can coexist with any block exposure, so level does not
identify structure. Part (b) is the converse discipline: one safety writer can
lie outside a particular nuisance span, giving $\chi_B=0$, while
overcompleteness forces nonzero total cross-talk somewhere in the system, and
equal-leverage frames are exactly the geometries that spread it evenly
(\cref{fig:globalfloor}), refining Welch-type floors to the cross-talk that
survives after every reader is optimized.

The global stage closes the arc at the capacity limit
(\cref{fig:globalfloor}, appendix): equal-leverage frames attain the
reader-optimized floor of \cref{thm:global}(b) at ratio $\GlobalEqualRatio$,
random unit frames average $\GlobalRandomOptimizedRatio$ of the floor while
their tied readers retain $\GlobalRandomTiedRatio$ times the optimized
cross-talk, and the benefit of reader optimization grows with leverage
unevenness (Spearman $\GlobalLeverageSpearman$).

The same aggregate viewpoint has a live counterpart in interpretability.
\citet{gorton2025adversarial} measure superposition by features per dimension
and report a near-perfect correlation with adversarial vulnerability across
toy models trained at different sparsities. \Cref{thm:global}(c) shows that
such an aggregate leaves the location of exposure free: a column permutation
holds the spectrum, rank, trace and frame potential fixed while moving the
block exposure $\rho_{\isafe,B}$ to any admissible value, and
\cref{sec:importance} shows when their correlation is nevertheless the outcome
to expect. The attainment side is classical frame theory rather than new here:
the aggregate floor is the coherence bound of \citet{welch1974lower}, and
equality for equal-norm tight frames is the frame-potential result of
\citet{benedetto2003finite}. What \cref{thm:global}(b) adds is that the same
constant survives after every reader is optimized, with equality exactly at
equal leverage. Read as an allocation, $\sum_i\ell_i=\rank(W)$ is the budget
that \citet{scherlis2022capacity} identify as the fractional dimension each
feature consumes; the leverage form makes each share dual to a readout
objective, so the budget acquires a price. The floor itself is the per-block story summed
over the dictionary: each row minimum in \cref{eq:globalfloor} is
$1/\ell_i-1=\chi_i^2$, so the reader-optimized total is exactly the aggregate
intrinsic exposure, and the comparison quantifies the split. Random frames
land within about two percent of the floor in aggregate while their tied
readers carry $\GlobalRandomTiedRatio$ times the optimized cross-talk,
roughly a third of it removable, with the gain of optimization tracking
leverage unevenness (Spearman $\GlobalLeverageSpearman$). The aggregate is
pinned near a rank-determined constant while its allocation across blocks,
the only thing a safety audit cares about, remains free.
\label{sec:importance}

The counterexample of \cref{thm:global}(c) is a construction. Whether the level
identifies exposure in frames that training actually produces is a separate,
empirical question, and the answer depends on one design variable: how
unequally the features matter.

The stage trains $\ImportanceModels$ overcomplete reconstruction models
($n=80$ features, $d=20$ dimensions, ReLU output, sparse nonnegative data) for
$\ImportanceSteps$ Adam steps at eight sparsities, under uniform importance
and under two geometrically decaying importance profiles, and attacks each with
a one-step gradient step at a fixed relative budget; $\ImportanceKept$ models
carry enough represented capacity to enter the statistics. Importance-driven
capacity allocation is the mechanism identified by
\citet{elhage2022toy,scherlis2022capacity}: important features are given
something closer to a private direction, unimportant ones share. What the
audit adds is the consequence for a readout's worst-case exposure.

Under uniform importance the trained frames sit on the tight-frame floor:
tightness stays in $[\ImportanceTauUniformMin,\ImportanceTauUniformMax]$ and
the mean Welch gap is $\ImportanceWelchGapUniform$. On that floor the level and
the aggregate interference are the same variable, with correlation
$\ImportanceAlignUniform$ between their logarithms, and the level duly predicts
attacked loss ($R^2=\ImportanceRsqLevelUniform$). No feature is privileged: the
interference of the first feature is $\ImportanceRhoRatioUniform$ times the
median feature's.

Decaying importance leaves the floor. Tightness reaches
$\ImportanceTauNonuniformMax$, the mean Welch gap rises to
$\ImportanceWelchGapNonuniform$, the log-log correlation between level and
aggregate interference falls to $\ImportanceAlignNonuniform$, and the most
important feature is protected: its interference falls to a median
$\ImportanceRhoRatioNonuniform$ of the median feature's, reaching
$\ImportanceRhoTopNonuniformMin$ in absolute terms at a level where other
features remain fully entangled. Pooled over both families, the level explains
$R^2=\ImportanceRsqLevelPooled$ of the variation in attacked loss and the
global worst-case interference explains $R^2=\ImportanceRsqStructurePooled$:
once importance is unequal, neither aggregate summarises the vulnerability,
because both average over an allocation that has become uneven.

This is the trained counterpart of \cref{thm:global}(c), and it locates the
regime in which the aggregate correlations reported in this literature hold.
They are a property of equal importance, not a law about superposition; and the
quantity that survives the change of regime is not an aggregate at all but the
per-block exposure of a declared readout, which is what the rest of the paper
measures. The models here are toy reconstruction networks under one attack, so
the claim is about which functional identifies vulnerability in that setting,
not about language models.

\begin{figure}[tbp]
\centering
\includegraphics[width=\linewidth]{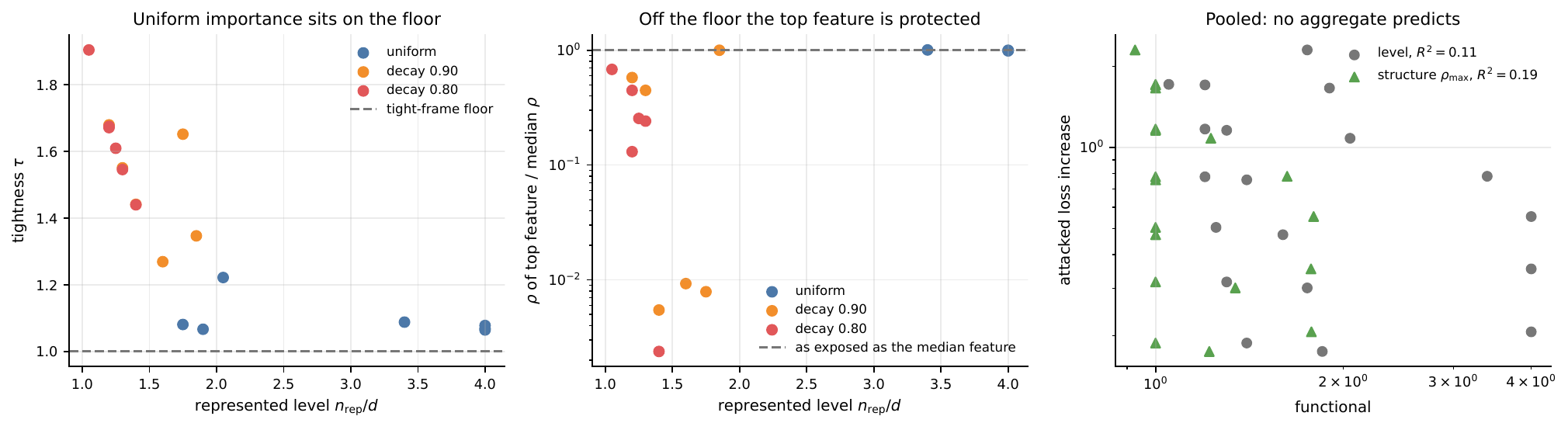}
\caption{Trained frames. Left: uniform importance holds tightness at the
tight-frame floor. Middle: with decaying importance the most important
feature's interference falls far below the median feature's. Right: pooled
across families, neither the level nor the global worst-case interference
tracks attacked loss.}
\label{fig:importance}
\end{figure}

\FloatBarrier
\section{The audit in practice}
\label{sec:recipe}

The workflow runs with or without a learned dictionary, and its order matters
more than any single quantity in it. Fibers are declared before outcomes are
examined, split by semantic intent so that realizations of one intent never
cross the training, validation and test sets. The score and its operating point
are fixed next: layer, reader, threshold and policy category, using the matrix
form \eqref{eq:multioutput} when a mechanism carries several categories at once.
The contrast geometry follows, as $W_B$ or the cross-Gram row where a dictionary
is available and as the hidden differences $d_k=h(x'_k)-h(x_k)$ with
$\widehat\Sigma_\Delta=N^{-1}\sum_kd_kd_k^\top$ where one is not. Reporting then
follows the declared budget, $\rho$ for a feature-coordinate ball, $\kappa$ for
a hidden-distance ball and $r^\top\widehat\Sigma_\Delta r$ for the observed
contrast distribution. Repairability comes next, from $\chi$ and $\norm{\rinv}$
with a dictionary and from the regularization path and the reader norm in the
covariance-only version. Regularization is selected on validation fibers under a
utility floor fixed in advance, and the reader, threshold and regularizer are
frozen before the single test evaluation. Where interventions are available, the
observed order swaps are compared with \eqref{eq:untiedcommutator}, and the
residual reports how far the linear control abstraction carries.

The geometry then names the work. Small current exposure calls for monitoring on
held-out fibers. High $\rho$ with small $\chi$ and moderate $\norm{\rinv}$ is a
reader that listens to more than it needs, so recalibration or a contrast
regularized reader settles it. The same $\rho$ and $\chi$ with a large
$\norm{\rinv}$ places the work on the regularized reader or on the
representation, since exact cancellation there is nominal rather than usable. A
large $\chi$ places it on the representation itself, through the encoder, an
adapter or dedicated capacity. A large order-swap residual points at
state-dependent or off-dictionary control behaviour, and a large leverage
imbalance points at how capacity is allocated across the dictionary rather than
at any single block.

Steps 3 to 5 apply per layer: running them at every depth, language, and
safety category produces a layer-resolved repairability profile showing where
the audit should attach and which layers admit a head-side repair at all. The
natural attachment point is the window where $\chi_\ell$ is near zero with
moderate conditioning; a scored layer outside that window puts the repair
budget on the representation rather than the head. Where that window sits is a
measurement. Layer choice for multilingual
steering has otherwise been made either heuristically or by an alignment and
separability criterion evaluated on downstream generation
\citep{alghussin2026multilingual}; $\chi_\ell$ with the invariant-reader norm
answers the prior question, whether a head-side repair exists at that depth. The
premise the audit rests on, that same-meaning inputs
share dictionary features across languages and scripts, is itself measurable
\citep{karne2026autointerp}, which is why the fibers are
audited externally rather than inferred from feature overlap.

The dictionary-free regularized reader is the closed form of
\cref{thm:ridge} with $W_BW_B^\top$ replaced by $\widehat\Sigma_\Delta$: a
single linear solve against the ridge-regularized contrast covariance,
renormalized so the calibration constraint holds exactly. For large hidden
dimensions the solve can use a low-rank SVD, conjugate gradients, or a
Woodbury identity rather than a dense inverse.
\subsection{The audit on a real encoder}
\label{sec:pilot}

The calculus above is established on synthetic geometry. Here it runs once on a
real encoder, in the covariance form that a single pooled layer supplies, to see
which regime a deployed score occupies.

The setting is small and fully specified. The representation is
the mean-pooled final layer of a public multilingual sentence encoder
($d=\PilotDimension$), read through a fixed inference session. The fibers are
$\PilotFibers$ professionally translated FLORES-200 sentences, so
same-meaning membership is external to the model and was fixed before any
activation was computed: $\PilotAuditedLanguages$ languages are audited and
$\PilotHeldOutLanguages$ are held back entirely. The score is a zero-shot
inner product with a fixed English query direction, which keeps the deployed
map linear in the representation. Fibers, not sentences, are split, so no
realization of one meaning appears in two splits.

The first measurement is the diagnosis. The audited contrast covariance,
estimated on $\PilotTrainFibers$ training fibers, has rank
$\PilotContrastRank$ at $99\%$ energy, and the deployed score direction lies
almost entirely inside that span: $\kappa=\PilotKappa$ of its unit norm.
Exact invariance is therefore available in principle, and useless in practice:
the minimum-norm invariant reader has norm $\PilotReaderNorm$, and applying it
raises mean absolute drift on untouched test fibers by
$\PilotProjectedReductionPct\%$ while its correlation with the original English
score falls to $\PilotProjectedUtility$. The exact fix does not fail because
the algebra is wrong; it fails because an estimated $\PilotContrastRank$
dimensional annihilator is fitted noise, which is the ill-conditioned regime
the triage is built to detect and the statistical failure a companion
manuscript characterizes \citep{ahnouch2026interpolation}.

The regularized reader of \cref{thm:ridge}, in the covariance form of the
recipe above, is the response the triage prescribes. Selecting the ridge on
validation fibers under a prespecified utility floor picks
$\lambda=\PilotSelectedRidgeRelative$ times the mean contrast eigenvalue, a
reader of norm $\PilotSelectedReaderNorm$ rather than $\PilotReaderNorm$. On
the untouched test split it lowers mean absolute drift from
$\PilotAuditedDeployedDrift$ to $\PilotAuditedSelectedDrift$ on audited
languages ($\PilotAuditedReductionPct\%$) and from
$\PilotHeldOutDeployedDrift$ to $\PilotHeldOutSelectedDrift$ on the languages
never audited ($\PilotHeldOutReductionPct\%$), while the English score
correlation stays at $\PilotSelectedUtility$.

The regime that occurs here is the conditioning one: a real safety direction sat
inside the language-contrast span, so the choice was between drift and reader
norm rather than between drift and exact invariance. The reduction that carries
to unaudited languages is of the same order as the audited one, which fits
language contrasts sharing much of their geometry. A single pooled layer
supports the covariance form of the calculus; the feature-coordinate constant
$\rho_{u,B}$, the intrinsic exposure $\chi_B$ against a declared block and the
order-swap residual are measured on the synthetic stacks, where read and write
directions are separate by construction. On one model, one layer, one score and
one language family, the diagnosis is computable and discriminating on real
activations.

\begin{figure}[tbp]
\centering
\includegraphics[width=\linewidth]{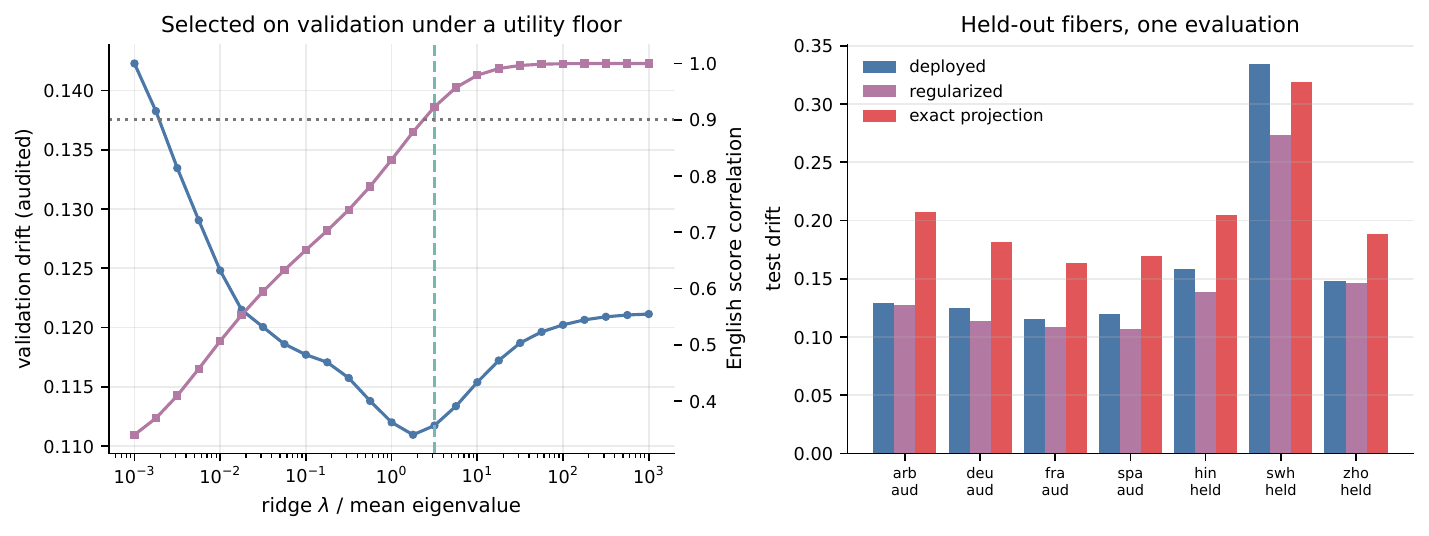}
\caption{The audit on a real encoder. Left: the validation frontier of the
regularized reader, with the utility floor and the selected ridge. Right: mean
absolute drift on untouched test fibers for the deployed direction, the
selected reader, and the exact projection, by language.}
\label{fig:pilot}
\end{figure}

\FloatBarrier
\section{Conclusion}

One symptom, a refusal that holds in English and fails in translation, has
three causes, and the audit gives each a number and an action. High current
exposure $\rho$ with low intrinsic exposure $\chi$ is a readout fault that
recalibration or a least-change projection removes. Low $\chi$ with a large
invariant-reader norm means the exact fix exists but is unstable, and a
regularized reader buys usable conditioning. High $\chi$ admits no readout
fix, and the repair budget belongs to the representation. Two limits frame
every such audit: aggregate spectra never reveal where exposure sits, so the
block must be measured directly, and total cross-talk has a floor that no
readout optimization crosses.

The triage applies wherever an explicit score reads a shared representation,
and \cref{sec:pilot} runs its covariance form once on a real multilingual
encoder, where the regime that occurred was the conditioning one. The next
measurement the calculus asks for is a sparse-autoencoder audit whose read and
write directions differ, where the order-swap residual is read from the stack
itself.

\subsection*{Scope}
The results describe one scored layer treated as a local linear model, which is
what keeps every quantity closed-form and checkable. The equivalence relation
arrives from outside the model, as a scientific and governance judgment about
which inputs deserve the same treatment. The synthetic study fixes dictionaries
and heads by construction and draws contrasts symmetrically, so the two language
directions are exchangeable and the three-regime diagnosis is checkable against
a known ground truth, with each rate anchored to one operating point. The block
$B$ is audited as given, and every setter obeys its algebraic definition, apart
from the untied stage, which injects nonlinear leakage to put the linear
interface under stress. Within that setting, $\rho_{u,B}$, $\kappa_{u,B}$,
$\sigma_\Delta$, $\chi_B$ and $\norm{\rinv}$ are the quantities an audit
reports, and \cref{sec:pilot} computes the subset that a single pooled layer
supplies on a real encoder.

\subsection*{Reproducibility Statement}
A single deterministic command regenerates every table, figure, and numerical
macro in this paper from local, explicitly seeded pseudorandom generators
(PCG64), on CPU in float64; each of the eleven stages runs in a fresh process.
Repeated runs on one machine reproduce the artifact manifest byte for byte.
Across BLAS builds three machine-epsilon identity checks and two sweep
correlations move in the last reported place or two, since those correlations
depend on discrete ridge selections that a last-place difference can flip; the
reported values are the released environment's. The real-encoder pilot of
\cref{sec:pilot} is a separate, optional script with its own pinned model
revision, single inference thread and committed outputs, so the manuscript
builds without downloading a model. Figure PDF metadata is pinned, a SHA-256 manifest
covers every generated artifact, and a provenance record accompanying the
release fixes the interpreter and library versions. A regression suite covers
the tie-aware AUC and threshold routines,
the relative-leverage identity, the exact-invariance regimes, the ridge KKT
condition, sample-level score matching, the drift identity, the explicit tied
and untied order-swap compositions, the sensitivity constants, the global
floor, and the macro/table generation contract. The code, data tables, and a
data manifest are released with the paper, and every text used in them is public
and benign.

\subsection*{Ethics Statement}
The work studies the safety of language models using public benign text
throughout. In the synthetic study ``language'' is an abstract feature block, and
the real-encoder pilot scores public professionally translated news sentences
against one benign English query, so the release contains no harmful corpus. The
claims concern a mathematical mechanism and the audit built on it. The work
involves no
human subjects or personal data. The proposed audit depends on externally
specified semantic equivalence, so real deployments should include domain
experts and native reviewers where language or culture is part of the fiber
construction.

\bibliographystyle{plainnat}
\bibliography{references}

\clearpage
\appendix

\section{Numerical protocol}

\emph{Matched benchmark generator.} The canonical safety writer is $e_0$.
With $\beta=0.22$ and $\gamma=\sqrt{1-\beta^2}$, every block column has the
form $\beta e_0+\gamma q_j$, so the deployed tied reader has exposure
$\beta\sqrt{|B|}$ in every family. The stable family uses orthonormal $q_j$;
the ill-conditioned family makes the final $q_j$ nearly dependent on the
others while retaining full column rank; the intrinsic family uses antipodal
pairs as in \cref{prop:matched}. Each geometry receives an independent Haar
rotation. For label $y\in\{-1,+1\}$ the central hidden state is
$h_0=0.78\,y\,w_s+\xi$ with $\xi\sim\mathcal N(0,0.11^2\Id)$, block
coefficients are $c\sim\mathcal N(0,0.72^2\Id)$, and endpoint residuals have
scale $0.10$; all isotropic noise is generated before the family-specific
rotation, which is what matches deployed scores sample by sample across
families.

\emph{Untied generator.} Each of the $\UntiedSystems$ systems draws unit
writers, calibrated readers $r_i=(w_i+0.35\,\zeta_i)/\ip{w_i+0.35\,\zeta_i}{w_i}$
with unit random $\zeta_i$, and a leakage direction orthogonal to each writer.
The actual setter adds the nonlinear term
$\nu_i(\delta_i^2+\tfrac12\delta_i\tanh(v_i^\top h))v_i$ to the linear untied
setter, with feature-specific strengths $\nu_i$; pairwise residuals are
estimated on $96$ calibration states, and the held-out task composes three
controls in forward and reverse schedules on $160$ new states and targets.

\section{Additional figures}

\begin{figure}[h]
\centering
\includegraphics[width=0.9\linewidth]{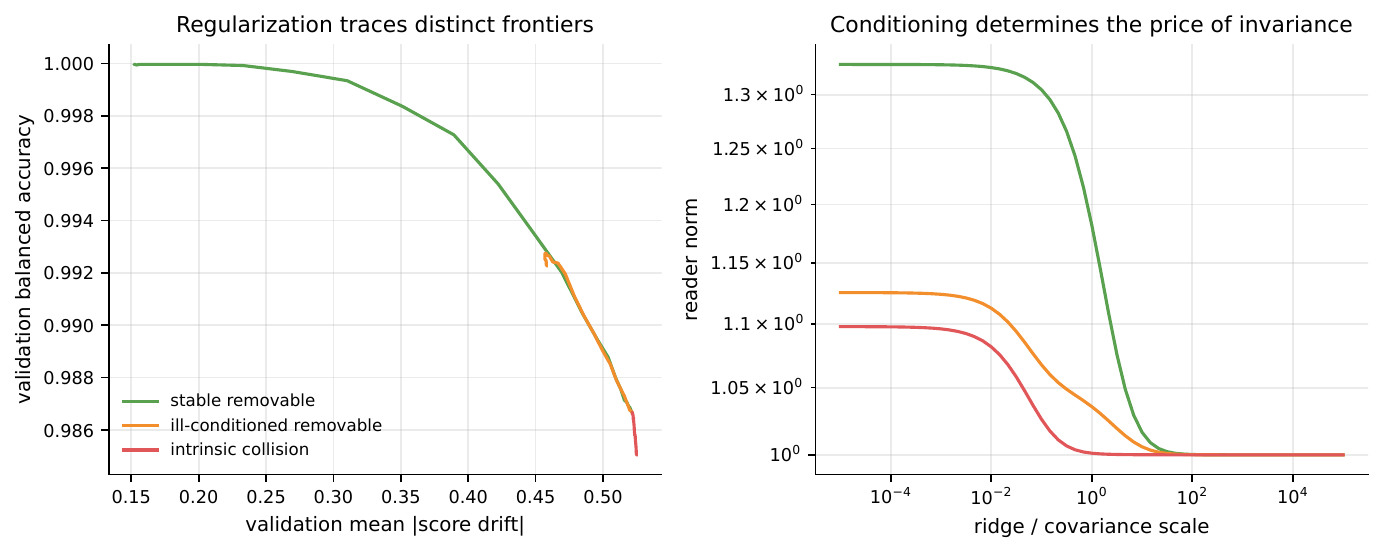}
\caption{Validation regularization frontiers behind \cref{tab:matched}. The
stable family approaches exact invariance at moderate reader norm; the
ill-conditioned family pays rapidly increasing norm, so validation selects a
partial repair; the intrinsic family returns to the deployed reader.}
\label{fig:frontiers}
\end{figure}

\begin{figure}[h]
\centering
\includegraphics[width=0.9\linewidth]{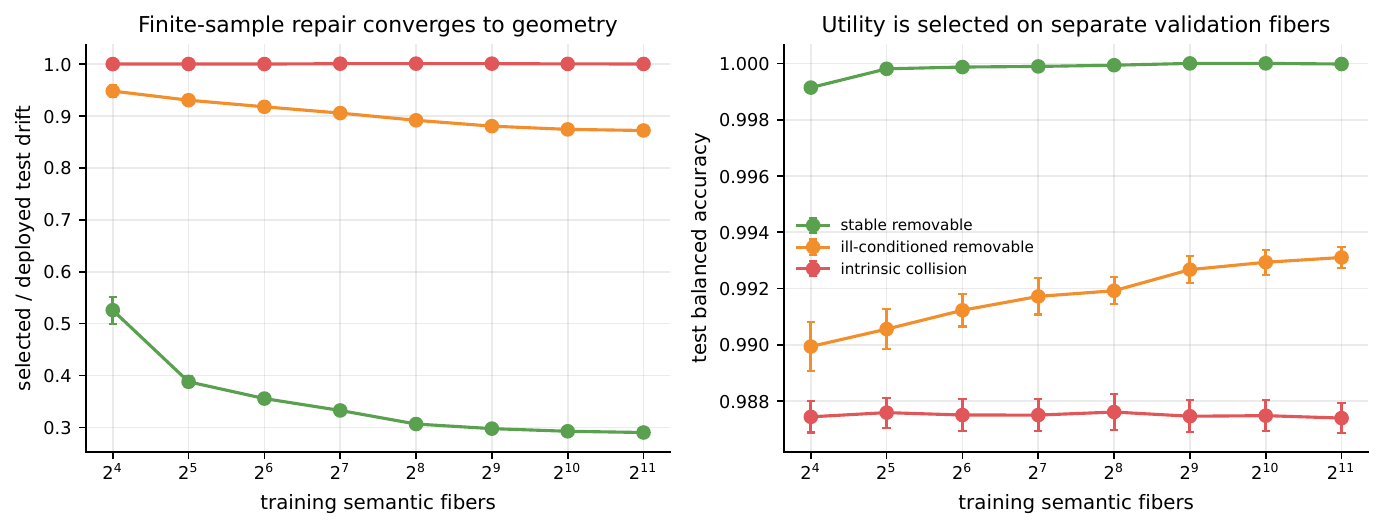}
\caption{Finite-sample learning curves: more audit pairs improve estimation,
not geometry. The intrinsic family's ratio is flat at
$\IntrinsicRatioMax$ regardless of sample size.}
\label{fig:learning}
\end{figure}

\begin{figure}[h]
\centering
\includegraphics[width=0.55\linewidth]{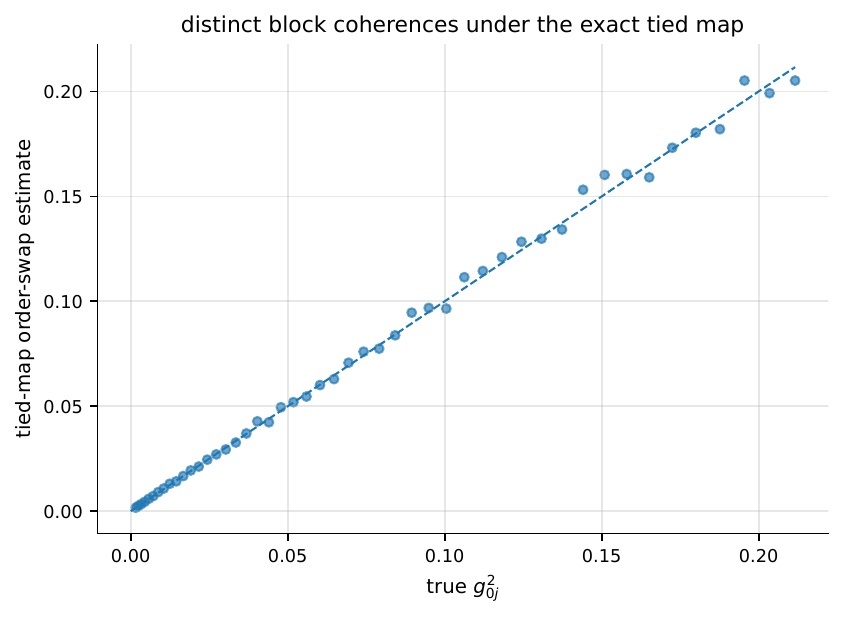}
\caption{Tied-map order-swap recovery of the $\CommDistinctPairs$ distinct
squared block coherences; a consistency check of the ideal tied map, matching
the closed form to $\CommIdentityError$.}
\label{fig:tiedcomm}
\end{figure}

\begin{figure}[h]
\centering
\includegraphics[width=0.9\linewidth]{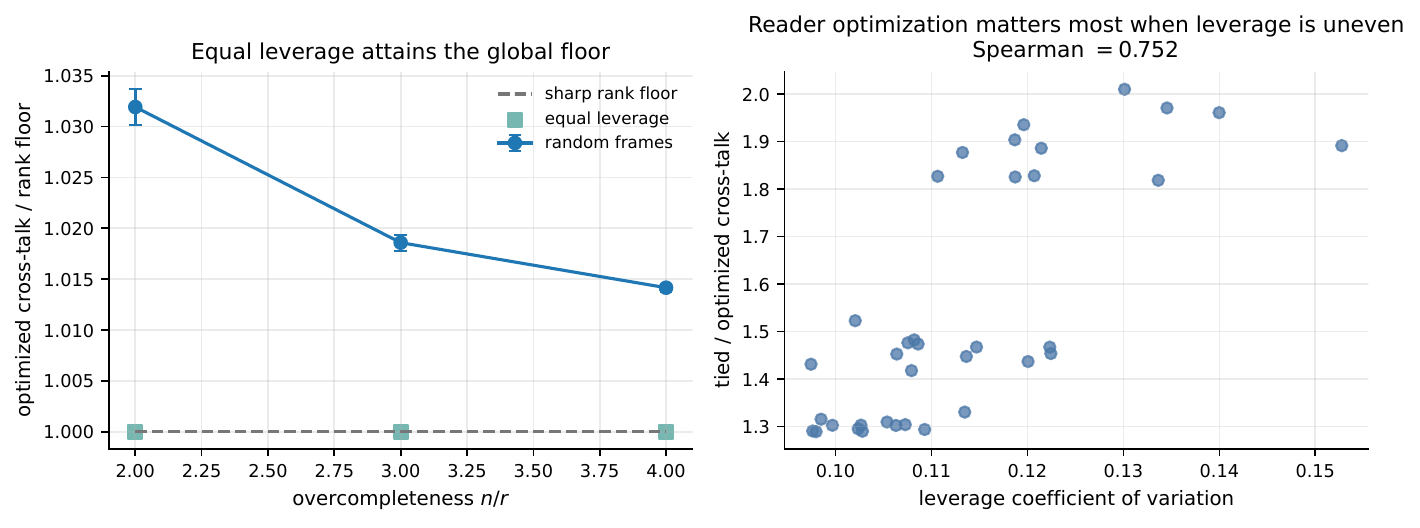}
\caption{The reader-optimized capacity floor of \cref{thm:global}(b).
Equal-leverage frames attain it exactly; the benefit of reader optimization
over tied readers grows with leverage unevenness.}
\label{fig:globalfloor}
\end{figure}

\end{document}